\documentclass[journal]{IEEEtran}

\usepackage{amsmath,amssymb,amsfonts}
\usepackage{graphicx}
\usepackage[hyphens]{url}
\usepackage{multirow}
\usepackage{makecell}
\usepackage{subfigure}
\usepackage{booktabs}
\usepackage{todonotes}
\usepackage{algorithm}
\usepackage{wasysym}
\usepackage{times}
\usepackage{graphicx}
\usepackage{grffile}
\usepackage{amssymb}
\usepackage{multirow}
\usepackage{wasysym}
\usepackage{caption}
\usepackage{url}
\usepackage{xcolor}
\usepackage{amsthm}
\usepackage{adjustbox}
\usepackage{eurosym}
\usepackage{array}
\usepackage{calc}
\usepackage{mathtools}
\usepackage{pifont}
\usepackage{algpseudocode}

\usepackage[hidelinks]{hyperref}
\hypersetup{
pdfauthor={Christian Rathgeb, Johannes Merkle, Johanna Scholz, Benjamin Tams, Vanessa Nesterowicz},
pdfsubject={Biometrics},
pdftitle={Deep Face Fuzzy Vault: Implementation and Performance},
pdfkeywords={Biometric Template Protection, Biometric Cryptosystem, Fuzzy Vault Scheme, Face Recognition},
bookmarks=true,
bookmarksopen=true,
bookmarksnumbered=true,
pagebackref=false
}

\ifCLASSOPTIONcompsoc
  \usepackage[nocompress]{cite}
\else
  \usepackage{cite}
\fi

\def\eg{\textit{e.g}. } \def\Eg{\textit{E.g}. }
\def\ie{\textit{i.e}. } 
\def\cf{\textit{c.f}. }

\def\wrt{w.r.t. } \def\etal{\textit{et al}. }

\newcommand{\hash}{H}

\newcommand{\cmark}{\ding{51}}%

\begin{document}

\title{Deep Face Fuzzy Vault:\\ Implementation and Performance}

\author{Christian~Rathgeb, Johannes Merkle, Johanna Scholz, Benjamin Tams, Vanessa Nesterowicz
\IEEEcompsocitemizethanks{\IEEEcompsocthanksitem The authors are with secunet Security Networks AG, Essen, Germany. C. Rathgeb is also with the Hochschule Darmstadt, Germany. J. Scholz and V. Nesterowicz are also with the Ruhr-Universität Bochum. \protect\\
E-mail: \{name\}.\{lastname\}@secunet.com
}
}

\maketitle
\thispagestyle{empty}
\vspace{-0.4cm}
\begin{abstract}

Biometric technologies, especially face recognition, have become an essential part of identity management systems worldwide. 
In deployments of biometrics, secure storage of biometric information is necessary in order to protect the users' privacy. In this context, biometric cryptosystems are designed to meet key requirements of biometric information protection enabling a privacy-preserving storage and comparison of biometric data, e.g. feature vectors extracted from facial images. Until now, biometric cryptosystems have hardly been applied to state-of-the-art biometric recognition systems utilizing deep convolutional neural networks.

This work investigates the application of a well-known biometric cryptosystem, i.e. the improved fuzzy vault scheme, to facial feature vectors extracted through deep convolutional neural networks. To this end, a  feature transformation method is introduced which maps fixed-length real-valued deep feature vectors to integer-valued feature sets. As part of said feature transformation, a detailed analysis of different feature quantisation and binarisation techniques is conducted.  At key binding, obtained feature sets are locked in an unlinkable improved fuzzy vault. For key retrieval, the efficiency of different polynomial reconstruction techniques is investigated. The proposed feature transformation method and template protection scheme are agnostic of the biometric characteristic and, thus, can be applied to virtually any biometric features computed by a deep neural network. In experiments, an unlinkable improved deep face fuzzy vault-based template protection scheme is constructed employing features extracted with a state-of-the-art deep convolutional neural network trained with the additive angular margin loss (ArcFace). For the best configuration, a false non-match rate below 1\% at a false match rate of 0.01\%, is achieved in cross-database experiments on the FERET and FRGCv2 face databases. On average, a security level of up to approximately 28 bits is obtained. This work presents an effective face-based fuzzy vault scheme providing privacy protection of facial reference data as well as digital key derivation from face.
\end{abstract}


\IEEEpeerreviewmaketitle

\section{Introduction}\label{sec:introduction}
\IEEEPARstart{F}{ace} recognition technologies are employed in many personal, commercial, and governmental identity management systems around the world. In a face recognition system, a \textit{reference face image} is captured at enrolment; the face is detected, pre-processed, and a feature vector is extracted which is stored as \textit{reference template}. At the time of authentication, a \textit{probe face image} is captured, processed in the same way, and compared against a reference template of a claimed identity (verification) or up to all stored reference templates (identification). For a long period of time, handcrafted feature extractors, \eg Local Binary Patterns \cite{Ahonen06} and Gabor filters \cite{SHEN2007553}, were predominately used. Said methods apply texture descriptors locally and aggregate extracted features into an overall face descriptor. A large variety of such systems has been proposed in the scientific literature, see \cite{Li2011,Liu2019}. In contrast, state-of-the-art face recognition technologies utilise deep learning and massive training datasets to learn rich and compact representations of faces \cite{Parkhi2015,Guo-DeepFaceSurvey-2019}. The recent developments in Deep Convolutional Neural Networks (DCNNs) have led to breakthrough advances in facial recognition accuracy, surpassing human-level performance \cite{Taigman14,Ranjan18a}.  Similar kinds of developments, e.g. deconvolutional neural networks, have shown impressive results for reconstructing face images from their corresponding embeddings in the latent space \cite{Mai19a}. This poses a severe security risk which necessitates the protection of stored deep face embeddings in order to prevent from misuse, e.g. identity fraud. In response, directives and regulations have already been stipulated by legislators.

Privacy regulations, \eg the General Data Protection Regulation (GDPR) \cite{EU-GDPR-2016}, generally classify biometric templates as sensitive data which requires protection. It is well-known that traditional encryption methods are unsuitable for protecting biometric data, due to the natural intra-class variance of biometric characteristics, in particular the face. More precisely, biometric variance prevents from a biometric comparison in the encrypted domain, \ie analogous to password hashing. Consequently, the use of conventional cryptographic methods would require a decryption of protected biometric data prior to the comparison. In contrast, \emph{biometric template protection} \cite{5396650,Rathgeb11e,BNandakumar15a} enables a comparison of biometric data in the encrypted domain and hence a permanent protection of biometric data.

Biometric template protection has been an active field of research for more than two decades. For comprehensive surveys on this topic the interested reader is referred to \cite{Rathgeb11e,BNandakumar15a}. Biometric template protection methods are commonly categorised as \emph{cancelable biometrics} and \emph{biometric cryptosystems}.  Cancelable biometrics employ transforms in signal or feature domain which enable a biometric comparison in the transformed (encrypted) domain \cite{Patel-CancelableBiometrics-2015}. In contrast, the majority of biometric cryptosystems binds a key to a biometric feature vector resulting in a protected template. Biometric comparison is then performed indirectly by verifying the correctness of a retrieved key \cite{BUludag04a}. That is, biometric cryptosystems further allow for the derivation of digital keys from protected biometric templates, \eg fuzzy commitment \cite{BJuels99a} and fuzzy vault scheme \cite{BJuels02a}. Alternatively, homomorphic encryption has frequently been suggested for biometric template protection \cite{Aguilar-Homomorphic-2013}. Homomorphic encryption makes it possible to compute operations in the encrypted domain which are functionally equivalent to those in the plaintext domain and thus enables the estimation of certain distances between protected biometric templates. The requirements on biometric template protection schemes are defined in ISO/IEC IS 24745 \cite{ISO11-TemplateProtection}:

\begin{itemize}
\item \textbf{Unlinkability}: the infeasibility of determining if two or more protected templates were derived from the same biometric instance, \eg face. By fulfilling this property, cross-matching across different databases is prevented. 
\item \textbf{Irreversibility}: the infeasibility of reconstructing the original biometric data given a protected template and its corresponding auxiliary data. With this property fulfilled, the privacy of the users' data is increased, and additionally the security of the system is increased against presentation and replay attacks.
\item \textbf{Renewability}: the possibility of revoking old protected templates and creating new ones from the same biometric instance and/or sample, \eg face image. With this property fulfilled, it is possible to revoke and reissue the templates in case the database is compromised, thereby preventing misuse.
\item \textbf{Performance preservation}: the requirement of the biometric performance not being significantly impaired by the protection scheme.
\end{itemize}

Table~\ref{tab:overview_btp} gives an overview of the aforementioned categories of biometric template protection and their properties \wrt the above criteria as well as key derivation and efficient biometric comparison. In contrast to homomorphic encryption, the vast majority of works on cancelable biometrics and biometric cryptosystems reports a performance gap between protected and original (unprotected) systems \cite{BNandakumar15a}. Cancelable biometrics usually employ a biometric comparator similar or equal to that of unprotected biometric systems. Thereby, cancelable biometrics are expected to maintain the comparison speed of the unprotected system which makes them also suitable for biometric identification \cite{Drozdowski-WorkloadSurvey-IET-2019}. Biometric cryptosystems may need more complex comparators. Similarly, homomorphic encryption usually requires higher computational effort. In contrast to cancelable biometrics and homomorphic encryption, biometric cryptosystems enable the binding and retrieval of digital keys.

\begin{table}[!t]
\centering
\caption{Properties of template protection categories.}\label{tab:overview_btp}\vspace{-0.1cm}
\resizebox{.495\textwidth}{!}{
\renewcommand*{\arraystretch}{1.2}
\begin{tabular}{|c|c|c|c|c|c|c|c|}
\hline
\begin{tabular}{@{}c@{}}\textbf{Template}\\ \textbf{protection}\\ \textbf{category}\end{tabular} & \rotatebox[origin=c]{90}{\textbf{Unlinkability}} & \rotatebox[origin=c]{90}{\textbf{\hspace{0.1 mm} Irreversibility \hspace{0.1 mm}}} & \rotatebox[origin=c]{90}{\textbf{Renewability}} & \rotatebox[origin=c]{90}{\begin{tabular}{@{}c@{}}\textbf{Performance}\\ \textbf{preservation}\end{tabular}} & \rotatebox[origin=c]{90}{\begin{tabular}{@{}c@{}}\textbf{Efficient}\\ \textbf{comparison}\end{tabular}}  &  \rotatebox[origin=c]{90}{\begin{tabular}{@{}c@{}}\textbf{Key}\\ \textbf{derivation}\end{tabular}} 
\\ \hline
\begin{tabular}{@{}c@{}}Cancelable\\ biometrics \end{tabular} & \quad\cmark \quad \quad & \quad\cmark \quad \quad & \quad\cmark \quad \quad & (\cmark) & \cmark &  \\\hline
\begin{tabular}{@{}c@{}}Biometric\\ cryptosystems \end{tabular}  & \cmark & \cmark & \cmark & (\cmark) & (\cmark) &  \cmark \\\hline
\begin{tabular}{@{}c@{}}Homomorphic \\ encryption\end{tabular}  & \cmark & \cmark & \cmark & \cmark & (\cmark) & \\\hline
 \end{tabular}
}
\end{table}

In the past decades, numerous biometric template protection schemes have been proposed for various biometric characteristics, including face, see subsection~\ref{sec:templateprotection}. This large amount of research notwithstanding, face-based biometric cryptosystems have received relatively little attention in biometric research. This may be explained by the limited biometric performance achieved by face recognition during the high time of research on biometric template protection (approximately in the early 2000s). Obviously, privacy protection and sizes of derived keys are theoretically upper bounded by the biometric performance of the underlying recognition system, in particular its false match rate \cite{Ballard08,Ignatenko09a}. As mentioned before, the biometric performance of face recognition has significantly improved in the recent past such that low false match rates are achieved at practical false non-match rates. This fact motivates the \mbox{(re-)investigation} of face-based biometric cryptosystems.

In this work, an unlinkable improved face fuzzy vault-based cryptosystem is proposed which enables the protection of deep face embeddings, hereafter referred to as \emph{deep face representations}, as well as key derivation thereof. In this context, the main contribution of this work are:
\begin{itemize}
	\item A biometric characteristic-agnostic feature transformation is introduced which transforms a real-valued feature vector to a set of integer features. This transformation is based on a three-stage process involving a feature quantisation, a feature binarisation, and a feature set mapping step. The proposed feature transformation method can be applied to any fixed-length real-valued feature vectors as commonly extracted by DCNNs.
	\item The proposal of an unlinkable improved fuzzy vault scheme adapted to deep face recognition. Required key binding and key retrieval processes are described in detail. Additionally, different polynomial reconstruction techniques are considered for key retrieval. The presented face fuzzy vault can be easily extended to a multi-biometric scheme using feature-level fusion, whereby a very high level of security can be achieved. 
	\item A comprehensive performance evaluation of combinations of different quantisation and binarisation methods in cross-database experiments on two publicly available face databases using an open-source deep face recognition system. In addition, a detailed performance evaluation of the proposed face-based fuzzy vault is given where multiple configurations of the proposed feature transformation are investigated. Moreover, various decoding strategies and their runtime are analysed for key retrieval and a comparison against published approaches is provided. Finally, the security of the proposed face fuzzy vault is analysed considering various attacks and a comparison against previous works is made.
\end{itemize}

The remainder of this work is organised as follows: section~\ref{sec:background} revisits related works. The proposed fuzzy vault scheme is described in detail in section~\ref{sec:facefvs} and experiments based on deep face representations are presented and discussed in section~\ref{sec:experiments}. Concluding remarks are given in section~\ref{sec:conclusion}.
\section{Related Work}
\label{sec:background}

Different research fields are directly related to this work, namely biometric feature type transformation, face-based template protection in general, and the application of the fuzzy vault scheme to (face) biometric data. The following subsection discusses works on feature type transformations for biometric template protection (subsection~\ref{sec:featuretransformation}). Subsequently, the most relevant works on face-based biometric template protection are briefly summarised (subsection~\ref{sec:templateprotection}). Afterwards, the fuzzy vault scheme is revisited in detail (subsection~\ref{sec:fvs}).

\subsection{Feature Type Transformation}
\label{sec:featuretransformation}
Common feature representations have been established for templates of different biometric characteristics, \eg minutiae sets for fingerprints. However, biometric template protection schemes, in particular biometric cryptosystems, require templates in a distinct type of feature representation, \eg fixed-length binary strings for the fuzzy commitment scheme. In order to make biometric templates compatible to a template protection scheme, a feature type transformation may be necessary \cite{Lim-BiometricFeatureTypeTransformation-IEEE-2015}. This is particularly the case for multi-biometric template protection where fusion should be performed at the feature level to achieve high security levels \cite{Merkle12a}.

Focusing on face, biometric templates frequently consist of fixed-length integer- or real-valued feature vectors, \eg aggregated descriptors of facial regions or deep face representations extracted by DCNNs. To protect facial data in template protection schemes which take binary bit strings as input, numerous binarisation techniques have been proposed \cite{Lim-BiometricFeatureTypeTransformation-IEEE-2015}. A recent benchmark of popular binarisation schemes was presented in \cite{Drozdowski-DeepFaceBinarisation-ICIP-2018}. Additionally, techniques for adapting the intra-class variance of face feature vectors to error correction capabilities of biometric cryptosystems have been proposed, \eg in \cite{Ao09}. Recently, researchers have suggested  to train \mbox{DCNNs} to generate compact binary strings, commonly referred to as Deep Hashing \cite{Xia14a,Liong15a}. Such techniques have already been adapted to obtain binary face representations for biometric template protection, \eg in \cite{Talreja21,Mai21}. To obtain feature sets from binary face templates, various researchers have proposed to divide binary vectors into small bit-chunks and convert those directly to their decimal representation, \eg in \cite{BNagar12a}.

\subsection{Face Template Protection}
\label{sec:templateprotection}

In 2001, Ratha \etal \cite{BRatha01a} proposed the first cancelable face recognition system using image warping to transform biometric data in the image domain. Another popular cancelable transformation of face images based on random convolution kernels was presented in \cite{Savvides04a}. In contrast to \cite{BRatha01a}, this approach employs a fundamentally reversible distortion of the biometric signal based on some random seed which later coined the term ``biometric salting''. The majority of published  cancelable face recognition schemes applies transformations in the feature domain \cite{Patel-CancelableBiometrics-2015}. Over the past years, numerous feature transformations have been proposed in order to construct face-based cancelable biometrics, \eg BioHashing \cite{Teoh06}, BioTokens \cite{Boult06}, and Bloom filters \cite{GomezBarrero2016}. Recently, feature transformations have been specifically designed for DCNN-based face recognition, \eg stable hash extraction \cite{Jindal18a} or random subnetwork selection \cite{Mai21}. Analyses of some popular cancelable face recognition systems have uncovered security gaps, \eg in  \cite{KONG20061359,BRINGER2017239,Ghammam20,Kirchgasser20b}, and already led or are expected to lead to (continuous) improvements of such schemes. 

Regarding biometric cryptosystems, the fuzzy commitment scheme \cite{BJuels99a} and the fuzzy vault scheme \cite{BJuels02a} represent widely used cryptographic primitives. Both schemes enable an error-tolerant protection of (biometric) data by binding them with a secret, \ie key. Binarised face feature vectors have been protected through the fuzzy commitment scheme in various scientific publications, \eg in \cite{Ao09,8739177,keller2020fuzzy,Mai21}. However, it was shown that the fuzzy commitment scheme can not effectively protect against correlation attacks \cite{Tams14}. In contrast, only a few works have employed the fuzzy vault scheme for face template protection (see subsection~\ref{sec:fvs}). It is worth mentioning that some template protection approaches combine concepts of cancelable biometrics with those of biometric cryptosystems resulting in hybrid schemes \cite{Rathgeb11e}.

For a long time, homomorphic encryption has been considered as impractical for biometric template protection due to its computational workload. However, in the last years, homomorphic encryption has been applied effectively to face recognition where practical processing times could be achieved on commodity hardware \cite{Boddeti-HE-2019,Drozdowski-HomomorphicIdentificationFace-BIOSIG-2019,engelsma2020hers}. Depending on the used homomorphic cryptosystem, different feature type transformations might be required \cite{Kolberg20a}.

More recently, so-called privacy enhancing face recognition has been proposed by various researchers, \eg in \cite{Morales20,Terhoerst20}. The common goal of these approaches is to train a DCNN for face recognition in a way that suppresses demographic information  within deep face representations, \eg sex or age. Thus, privacy enhancing face recognition partially fulfills requirements of biometric template protection.

\begin{figure*}[!t]
\vspace{-0.0cm}
\centering
\includegraphics[width=\textwidth]{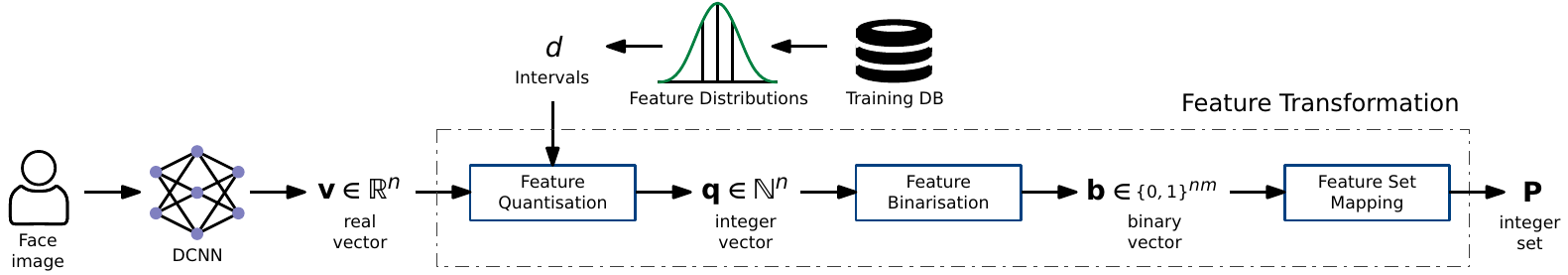}\vspace{-0.1cm}
\caption{Overview of the proposed feature transformation process: firstly, a real-valued feature vector is extarcted from a face image using a DCNN; based on a training step, the feature vector is quantised to an integer vector which is then binarised; lastly, an integer set is obtained from the binary vector through the feature set mapping.}\label{fig:featuretrans}\vspace{-0.2cm}
\end{figure*}

\subsection{Fuzzy Vault Scheme}
\label{sec:fvs}

The fuzzy vault scheme was introduced by Juels and Sudan \cite{BJuels02a,bib:JuelsSudan2006} and enables protection and error-tolerant verification with feature sets. It was suggested for the protection of fingerprint minutiae sets in \cite{bib:ClancyKiyavashLin2003}. Building upon this preliminary analysis, a series of implementations for minutiae-based fingerprint fuzzy vaults was proposed \cite{bib:NandakumarJainPankanti2007,bib:Nagar2010}. A useful guide for constructing a fuzzy vault scheme is provided in \cite{DBLP:conf/ivcnz/KrivokucaAS12}. Different security analyses have found that the original fuzzy vault scheme is vulnerable to a certain kind of linkage attack, \ie the correlation attack \cite{bib:ScheirerBoult2007,bib:KholmatovYanikoglu2008}. This conflicts with the above mentioned requirement of unlinkability as well as irreversibility. Dodis \etal \cite{bib:DodisEtAl2008} presented an improved version of the fuzzy vault scheme which generates much smaller records. It was later shown that that the aforementioned linkage attack can be effectively prevented in the improved fuzzy vault scheme \cite{bib:MerkleTams2013,bib:TamsMihailescuMunk2015}. 

Until now, the fuzzy vault scheme has been applied to various physiological as well as behavioural biometric characteristics, \eg iris \cite{BLee07b} and online signatures \cite{8954723}. Additionally, fuzzy vault schemes ultilising multi-biometric fuzzy vaults have been presented, \eg in \cite{bib:Tams2015,Rathgeb-ImprovedMultiBiometrics-EURASIP-2016}. 

Regarding the face, only a few early works applied the fuzzy vault scheme. Within thoses schemes, mostly handcrafted feature vectors are extracted and different feature type transformations are employed to obtain feature sets. Moreover, various decoding strategies were used.  Most published approaches were evaluated on rather small databases, \eg in \cite{Feng06a,Wang07}. Further, the majority of schemes reported impractical performance rates in terms of recognition accuracy in particular false match rates, \eg in  \cite{Wu10a}. High false match rates make those schemes vulnerable to false accept attacks which were rarely considered. The false accept security which is derived from a system's false match rate provides a good approximation of the actual security. In contrast, some works report brute-force security, \eg  41 bits in \cite{Feng06a}, which is usually a clear overestimate of the actual security considering corresponding recognition rates.  The work in \cite{BNagar12a} represents a notable exception, reporting a false accept security for a face-based fuzzy vault. Obviously,  most published approaches on face-based fuzzy vault systems did not employ DCNN-based face recognition which represent the current state-of-the-art, \cite{Dong21} representing a notable exception. A more detailed overview face-based fuzzy vault schemes and reported biometric performance as well as security rates is provided in subsection~\ref{sec:comprelated}.

For constructing a fuzzy vault-based cryptosystem, a practical decoding strategy is needed. To this end, a Reed-Solomon decoder \cite{bib:Gao2002} has been proposed in the original fuzzy vault scheme \cite{BJuels02a,bib:JuelsSudan2006}. In \cite{bib:NandakumarJainPankanti2007}, the repeated use of a Lagrange-based decoder has been suggested  and adopted for other implementations \cite{bib:Nagar2010,bib:TamsMihailescuMunk2015,Dong21}.  A reasonable trade-off between decoding time and verification performance can be achieved using a Guruswami-Sudan-based decoder \cite{bib:GuruswamiSudan1998}. In this work, these strategies are considered for key retrieval (see subsection~\ref{sec:bindingretrieval}).

\section{Fuzzy Vault for Deep Representations}
\label{sec:facefvs}
DCNNs are usually trained using \emph{differentiable} loss functions, \eg Euclidean distance. Consequently, deep face representations are represented as fixed-length real-valued vector $\mathbf{v}  \in \mathbb{R}^n$. This particularly applies to state-of-the-art face recognition systems. In the feature transformation step of the proposed system,  such fixed-length real-valued vectors are transformed to integer-valued feature sets (subsection~\ref{sec:featuretransform}). Subsequently, key binding and retrieval is performed in an unlinkable improved fuzzy vault scheme (subsection~\ref{sec:bindingretrieval}).

\subsection{Feature Transformation}
\label{sec:featuretransform}

An overview of the feature transformation process is shown in figure~\ref{fig:featuretrans}. While this method is designed for deep (face) representations, its basic principle could be applied to any type of fixed-length real-valued feature representation including those extracted by hand-crafted feature extractors, \eg texture descriptors. However, investigations based on such hand-crafted feature extractors are out of scope in this work, since these do not represent the state-of-the-art in face recognition. The applied  feature transformation process comprises the following three main steps: 

\subsubsection{Feature Quantisation}

In the feature quantisation step, a real-valued feature vector $\mathbf{v}= (v_i)^n_{i=1}, v_i \in \mathbb{R}$ is mapped to a quantised integer-valued feature vector $\mathbf{q}= (q_i)^n_{i=1}, q_i \in \mathbb{N}_0$ of same size. For this purpose, the probability densities of all $n$ feature elements are estimated. Based on its obtained probability density, the feature space of each feature element is then divided into $d=2^x$ integer-labelled intervals. Each element of the feature vector is then mapped to an integer number representing the corresponding interval on its support. Two quantisation schemes are applied:

\begin{itemize}
\item \textbf{Equal-probable intervals}: the feature space is divided into intervals containing equal population probability mass.
\item \textbf{Equal-size intervals}: the feature space is divided into intervals of equal size.
\end{itemize}

An example of both quantisation schemes is illustrated in figure~\ref{fig:bfeatureeconding}. The division of feature spaces of feature elements into intervals is determined based on a training database. Note that equal-size intervals can also be estimated directly given the feature space ranges.

\subsubsection{Feature Binarisation}\label{sec:binary}

In the binarisation step, the quantised feature vector $\mathbf{q}$  is mapped to a binary feature vector $\mathbf{b}\in  (b_i)^{nm}_{i=1}, b_i \in \{0,1\}$. Precisely, each quantised feature element $q_i$ (represented as integer) is mapped to a  binary string $b_i$ of length $m$. Subsequently, all binary strings are concatenated to produce the final binary feature representation of size $nm$. The dissimilarity of two such templates can be then computed using the Hamming distance. The following binarisation schemes are considered:

\begin{figure}[!t]
\vspace{-0.0cm}
\centering
\subfigure[equal-probable]{\includegraphics[width=0.24\textwidth]{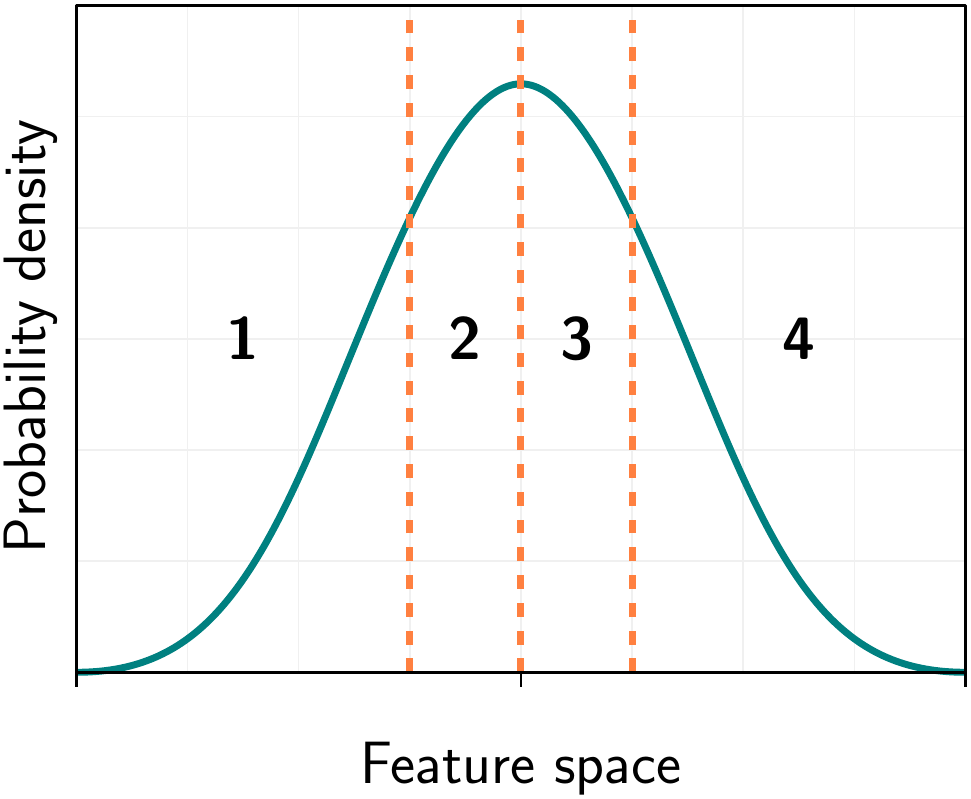}}\hfil
\subfigure[equal-size]{\includegraphics[width=0.24\textwidth]{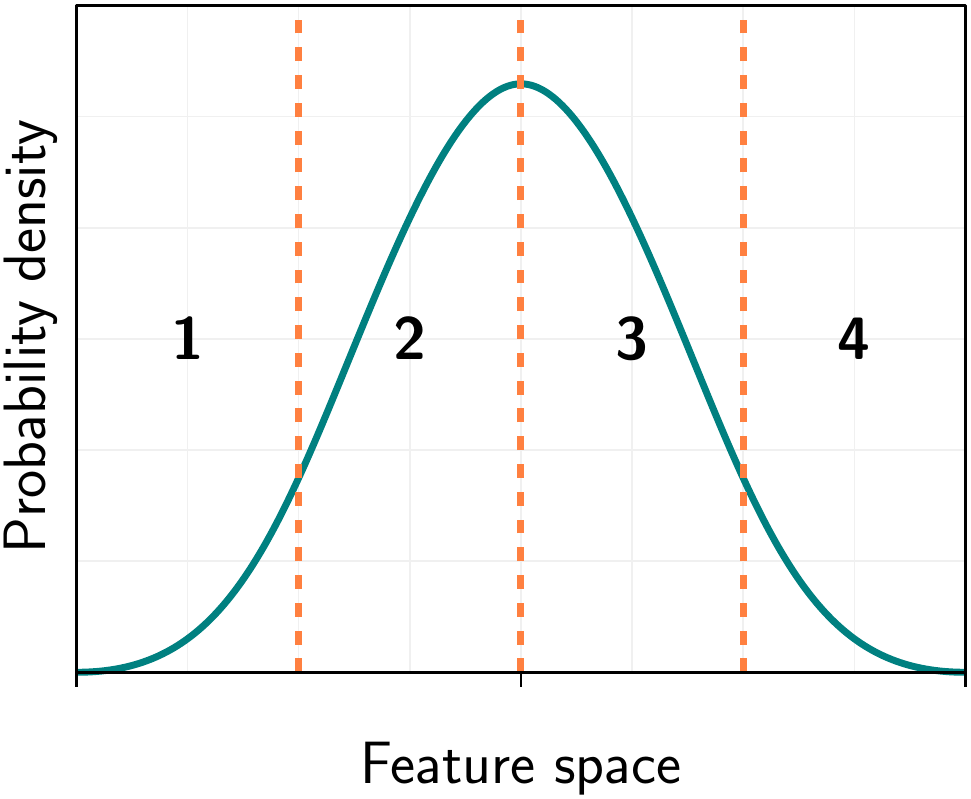}}
\caption{Example of quantisation of feature space in four intervals. }\label{fig:bfeatureeconding}\vspace{-0.2cm}
\end{figure}

\begin{itemize}
\item \textbf{Boolean}: The feature spaces are quantised into $d=2$ sub-spaces (\ie the resulting binary string is a single 0 or 1). In this simple scheme, the size of the quantised feature vector is maintained, \ie $m=1$.
\item \textbf{DBR} (Direct Binary Representation): In this scheme, the quantised feature elements are converted directly into their base-2 (binary) representations. The resulting binary vector is of size $nm$ with $m=\log_2(d)$.
\item \textbf{BRGC} (Binary Reflected Gray Code \cite{Gray-Patent-1953}): The binarisation is done in a way that the Hamming distance between codewords resulting from successive decimal values is always 1. The size of the binary vector is equal to that of  DBR.
\item \textbf{LSSC} (Linearly Separable Subcode \cite{Lim-LinearlySeparableSubcode-PAMI-2013}): A more recent approach, in which the distances between two binary values are equal to the $L1$ norm between the corresponding quantised values.  Compared to the previous schemes the size of the binary feature vector is significantly larger with $m=d-1$.
\item \textbf{One-hot}: In this scheme, the length of the binary representation is equal to that of the used intervals, \ie $m=d$. In each binarised value, only one bit is set to 1 which corresponds to the interval index resulting from the quantisation step (one-hot encoding). When applied to all feature elements, this results in a sparse binary feature vector. Note that the Hamming distance of such one-hot binary vectors is exactly twice that of the quantised feature vectors which is 1 for differing feature elements regardless of their distance. 
\end{itemize}

\begin{figure}[!t]
\vspace{-0.0cm}
\centering
\includegraphics[width=0.425\textwidth]{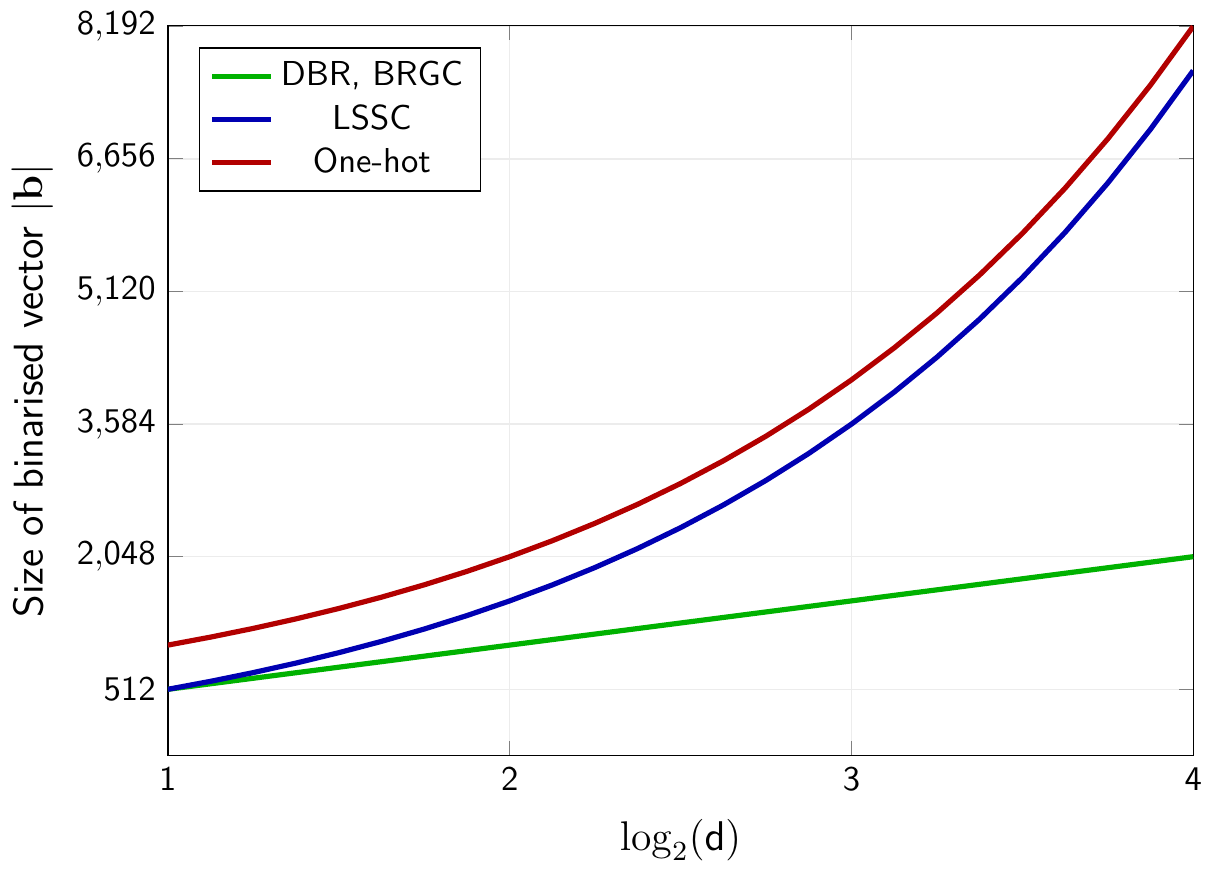}
\caption{Relation between used number of intervals and sizes of the binarised feature vectors for a quantised feature vector of size 512. }\label{fig:tempate_sizes}\vspace{-0.4cm}
\end{figure}

\begin{table}[!t]
\centering
\caption{Binarisation methods with four intervals.}
\label{table:encoding}
\resizebox{.4\textwidth}{!}{
\renewcommand*{\arraystretch}{1.2}
\begin{tabular}{|c|c|c|c|c|c|}
\hline
\multirowcell{2}{\textbf{Quantisation} \\ \textbf{Interval}} & \multicolumn{5}{c|}{\textbf{Method}} \\ 
& \textbf{Boolean} & \textbf{DBR} & \textbf{BRGC} & \textbf{LSSC} & \textbf{One-hot} \\ \hline
1 & 0 & 00 & 00 & 000 & 0001 \\\hline
2 & 1 & 01 & 01 & 001 & 0010 \\\hline
3 & -- & 10 & 11 & 011 & 0100 \\\hline
4 & -- & 11 & 10 & 111 & 1000 \\ \hline
\end{tabular}
}
\end{table}
Table \ref{table:encoding} shows an example for the binarisation methods described above with four intervals. A decrease in entropy density can be observed for the LSSC and the one-hot binarisation methods since theoretically only $2^{n \log_2(d)}$ different binary feature vectors can be generated for all binarisation methods. Intuitively, there exists a trade-off between the ability to obtain better separation, representation sparsity, and the required length of the binary vector. For a quantised vector of size 512, the relation between the number of intervals and the corresponding binarised vector  is plotted for the different binarisation methods in figure~\ref{fig:tempate_sizes}.

\subsubsection{Feature Set Mapping} In the last step of the feature transformation process, the binary feature vector $\mathbf{b}$ is mapped to a feature set $\mathbf{P}$. This feature set consists of all indexes of 1s in the binary vector,
\ie $\mathbf{P}=\{i | b_i=1\}$. The size of the feature set is equal to the Hamming weight of the binary vector, $|\mathbf{P}| = \mathit{HW}(\mathbf{b})$. This mapping of binary features to feature sets is different from those proposed in published works which usually map binary chunks to their decimal representation, \eg in \cite{BNagar12a}. In contrast, the proposed mapping is less sensitive to single bit flips and is therefore expected to obtain higher biometric performance in a fuzzy vault scheme. 

\subsection{Key Binding and Retrieval}
\label{sec:bindingretrieval}

\begin{figure}[!t]
\vspace{-0.0cm}
\centering
\subfigure[key binding]{\includegraphics[width=0.49\textwidth]{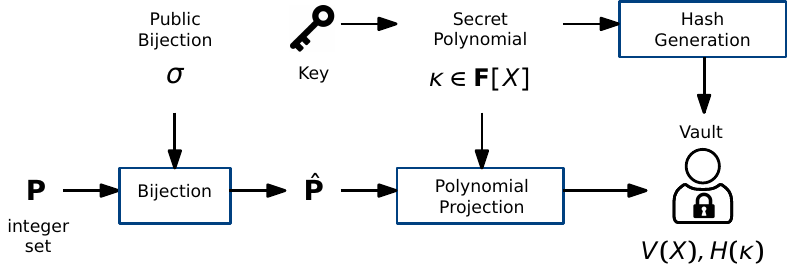}}\hfil
\subfigure[key retrieval]{\includegraphics[width=0.49\textwidth]{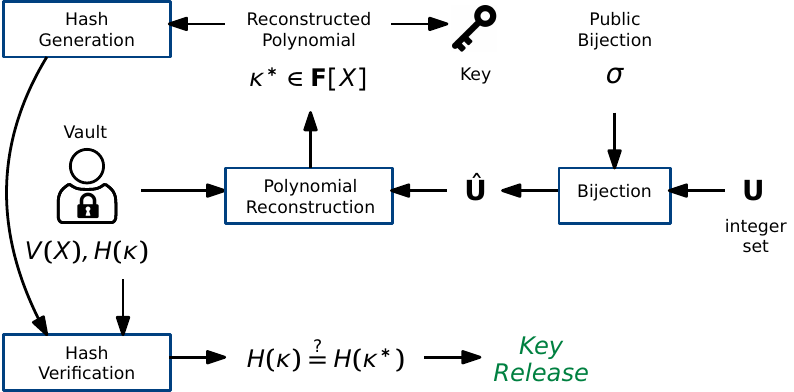}}
\caption{Overview of key binding and key retrieval: during key binding a public bijection is firstly applied to the integer set; subsequently, the resulting integer is projected onto a secret polynomial defined by a key; in addition, a hash of the key is stored. At key retrieval, the public public bijection is applied to another integer set and based on the resulting integer set the polynomial reconstruction is performed; finally the correctness of the retrieved key is validated by comparing its hash value to the stored one.}\label{fig:binding-retrieval}\vspace{-0.2cm}
\end{figure}

The key binding (enrolment) and retrieval (verification) processes are illustrated in figure~\ref{fig:binding-retrieval}. In the first step of the binding process, a secret polynomial $\kappa\in{\bf F}[X]$ of degree smaller than $k$ is chosen and the hash $\hash(\kappa)$ is stored. A record-specific but public bijection $\sigma:{\bf F}\rightarrow{\bf F}$ is applied to the feature set $\mathbf{P}$,  in order to re-map the elements of $\mathbf{P}$, $\hat{\mathbf{P}}=\sigma(\mathbf{P})=\{\sigma(v)|v\in\mathbf{P}\}$. To avoid additional data storage, it is suggested to use $\hash(\kappa)$ as seed to generate $\sigma$. This first step is performed as a countermeasure to the attack proposed in \cite{bib:MerkleTams2013}, see subsection~\ref{sec:runtime}. The application of a public bijection does not affect the biometric performance of the fuzzy vault scheme. More presicely, due to the properties of bijective functions (one-to-one correspondence) identical feature elements that occur in different feature sets will match before and after the application of the bijection. The same holds for non-matching feature elements.

The next step is performed based on the improved fuzzy vault scheme \cite{bib:DodisEtAl2008}. 
Feature elements are encoded by a monic polynomial of degree $t=|\hat{\mathbf{P}}|$. The features in $\hat{\mathbf{P}}$ are interpreted as elements of a finite field ${\bf F}$, $|{\bf F}|=\rho$, and bound to the secret polynomial $\kappa$ by estimating $V(X)=\kappa(X)+\prod_{v\in\hat{\mathbf{P}}}(X-v)$. The pair $(V(X),\hash(\kappa))$ is the final vault record. The elements of  $\hat{\mathbf{P}}$ can be represented with exactly $\log_2(nm)$ bits. The size of the vault increases with $t$, \ie the number of elements in the feature set $\hat{\mathbf{P}}$. This means, larger feature sets lead to an increased storage requirement. The maximum amount of feature elements possible is $nm$ which represents the upper bound for $t$. Consequently, the maximum size of the vault is upper bounded by $nm \log_2(nm)$ bits. 

At key retrieval, a probe feature set $\mathbf{U} \subset \mathbf{F}$ is computed. By
evaluating the polynomial $V(X)$ on its elements, a set of pairs  $\{\left(x, V(x)\right)| x \in
\mathbf{U}\}$ is obtained. Since $V(x) = \kappa(x)$ for $x\in \hat{\mathbf{P}}$, the pairs $\left(x, V(x)\right)$
with $x\in \mathbf{U} \cap \hat{\mathbf{P}}$ lie on the function curve of the secret polynomial $\kappa(X)$; these pairs
are referred to as genuine. If the number $\omega$ of genuine points is at least $k$, it is possible to reconstruct the polynomial $\kappa$ from ${\bf U}$. The correctness of $\kappa$ can be verified, \eg by using the hash value $\hash(\kappa)$. For the used feature extractor and the proposed feature transformation the unlocking set is expected to be sufficiently large to successfully recover $\kappa$, see section~\ref{sec:experiments}. 

The following polynomial reconstruction strategies are considered in this work:

\subsubsection{Iterated Lagrange Strategy} \label{sec:ils}
To reconstruct $\kappa$ from ${\bf U}$, $k$ pairs are selected from ${\bf U}$ and the unique polynomial $\kappa^*$ of degree smaller than $k$ that interpolates them is estimated. If all $k$ selected pairs are genuine, then $\kappa^*=\kappa$ which can be verified by observing $\hash(\kappa^*)=\hash(\kappa)$. If not all $k$ selected pairs are genuine, then most likely $\hash(\kappa^*)\neq\hash(\kappa)$ and the procedure is repeated until $\hash(\kappa^*)=\hash(\kappa)$. This procedure is guaranteed to, eventually, reconstruct the secret polynomial $\kappa$ if $\omega\geq k$. For a single step the success probability  is equal to
\begin{equation}
p_L(u,\omega,k)=\binom{\omega}{k}\cdot\binom{u}{k}^{-1}.
\end{equation}
Depending on the parameters $u$, $k$, and $\omega$, the iterated Lagrange strategy can become too costly and hence impractical as it may require a huge amount of iterations before $\kappa$ is recovered.

\subsubsection{Reed-Solomon Strategy}
Alternatively, a Reed-Solomon decoder, \eg see \cite{bib:Gao2002}, is capable of recovering $\kappa$ from ${\bf U}$ efficiently (by means of deterministic polynomial time) in case $\omega\geq(u+k)/2$. However, this class of algorithms will fail to recover $\kappa$ from ${\bf U}$ for $\omega<(u+k)/2$. To obtain a decoding mechanism which can deal with these cases, it is suggested to randomly select a $c$-sized subset ${\bf U}_0\subset{\bf U}$  where $|{\bf U}|\geq c\geq k$. Subsequently, the Reed-Solomon decoder can be applied to ${\bf U}_0$, and, if successfully revealing $\kappa^*$ with $\hash(\kappa^*)=\hash(\kappa)$, output the recovered polynomial; otherwise, this procedure is repeated until predefined number of iterations is reached. This procedure will succeed eventually if $\omega\geq(c+k)/2$ which improves upon the bound $\omega\geq(u+k)/2$ since $c\leq u$. The success probability for a single step is equal to
\begin{equation} \label{eq:pRS}
p_{RS}(u,\omega,k,c)=\binom{u}{c}^{-1}\sum_{j=\lceil(c+k)/2\rceil}^{\min(\omega,c)}\binom{\omega}{j}\cdot\binom{u-\omega}{c-j}.
\end{equation}

The Reed-Solomon decoding strategy is expected to significantly improve upon the iterated Lagrange method while it may still be too costly for a practical implementation. It is noteworthy that a Reed-Solomon decoder can be viewed as a special case of a Guruswami-Sudan decoder.
 
\subsubsection{Guruswami-Sudan Strategy}
By employing a Guruswami-Sudan list decoder \cite{bib:GuruswamiSudan1998}, the bound $\omega\geq(u+k)/2$ can be significantly improved. Provided that $\omega>\sqrt{u\cdot(k-1)}$, this algorithm can potentially recover $\kappa$\footnote{As a list decoder, the Guruswami-Sudan algorithm returns a list of candidate polynomials. From these, the correct can be easily determined by checking its hash value.}. While the Guruswami-Sudan decoder can be time-consuming in practice if one aims at recovering up to $u - \sqrt{u \cdot  (k-1)}$ errors, the computational efficiency can be traded-off against the number of correctable errors by an additional parameter referred to
as multiplicity. This algorithm may represent a significant improvement compared to a Reed-Solomon decoder.

An Guruswami-Sudan strategy will recover $\kappa$ from ${\bf U}$ in a single step with probability
\begin{equation}
p_{GS}(u,\omega,k,c)=\binom{u}{c}^{-1}\sum_{j=\lceil \sqrt{c \cdot (k-1)} \rceil}^{\min(\omega,c)}\binom{\omega}{j}\cdot\binom{u-\omega}{c-j}.
\end{equation}
This strategy is expected to outperform the iterated Lagrange as well as the Reed-Solomon strategies. 

The Guruswami-Sudan decoding method could be optimised by iteratively increasing the multiplicity until $\kappa$ is successfully recovered or a maximum number of iterations is reached. Moreover, it can be combined with one of the aforementioned decoding strategies as suggested in \cite{bib:Tams2015}. However, these optimizations are not deployed in our implementation.

\section{Application to Face and Evaluation}
\label{sec:experiments}
The following subsection describes the experimental setup for applying the proposed fuzzy vault to deep face representations (subsection~\ref{sec:setup}). Subsequently, the performance of the different variants of quantisation and binarisation methods are evaluated in a first experiment (subsection~\ref{sec:binmeth}). In the second experiment, the proposed feature transformation is applied using the best performing quantisation and binarisation schemes and the biometric performance of the deep face fuzzy vault scheme is estimated using the different decoding strategies (subsection~\ref{sec:fcsdec}). In the last experiment, a corresponding security and runtime analysis is presented (subsection~\ref{sec:runtime}). Finally, the proposed system is compared against other works (subsection~\ref{sec:comprelated}).

\subsection{Software and Databases}
\label{sec:setup}
For the extraction of deep face representations, the original implementation of the widely used ArcFace approach \cite{Deng19} is employed. This DCNN has been trained using the additive angular margin loss 
and achieves competitive recognition performance on various challenging datasets. In this work, we used the pre-trained model LResNet100E-IR,ArcFace@ms1m-refine-v2 published in the model zoo of the original ArcFace implementation\footnote{\url{https://github.com/deepinsight/insightface}}. This model takes input images of size 112$\times$112 pixels and extracts deep face representations of 512 floats. The ArcFace algorithm is a widely used DCNN that has been shown to obtain competitive biometric performance on many challenging databases. Different DCNN-based face recognition systems which have been proposed more recently may outperform the used ArcFace system. However, for the constrained scenario of this work, the ArcFace algorithm obtains very good recognition performance (as will be shown in subsequent subsections). For the image preprocessing (alignment, cropping and scaling), we closely followed the ArcFace implementation, but replaced the MTCNN face detector by the face and landmark detection of the dlib library \cite{King2009}.

\begin{figure}[t]%
	\centering
	\subfigure[FERET references]{
		\includegraphics[width=0.22\linewidth]{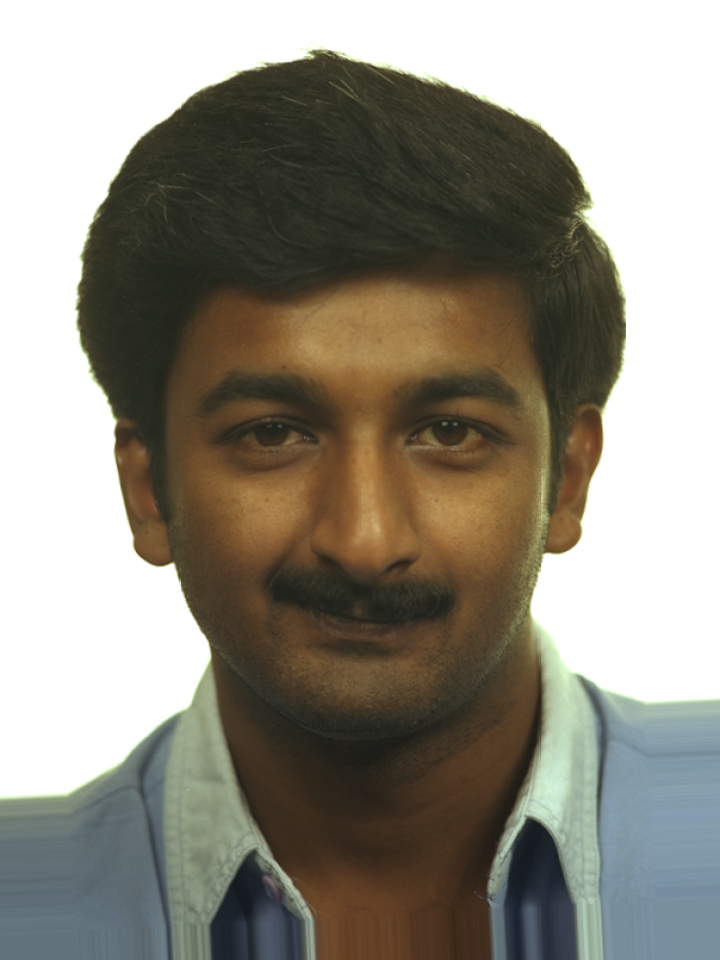}	
		\includegraphics[width=0.22\linewidth]{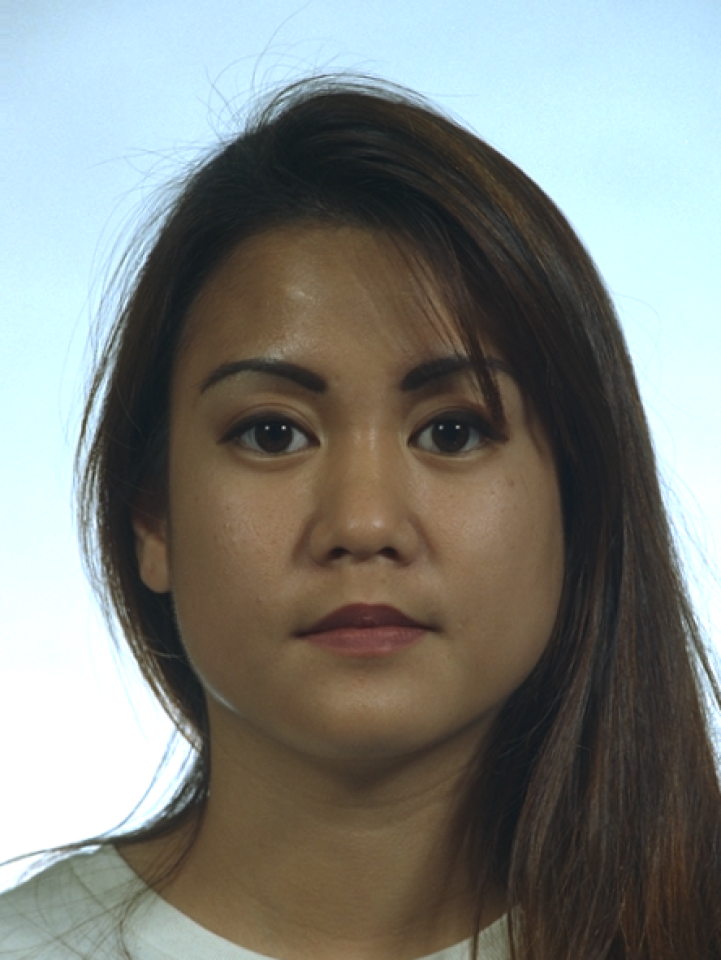}		
	}
	\subfigure[FRGCv2 references]{
		\includegraphics[width=0.22\linewidth]{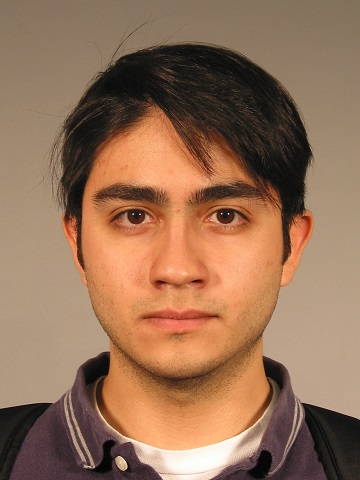}		
		\includegraphics[width=0.22\linewidth]{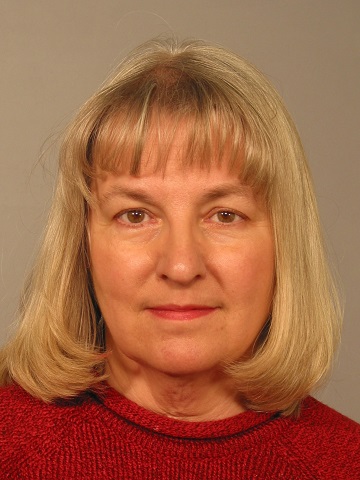}		
	}\\\vspace{-0.2cm}
	\subfigure[FERET probes]{
		\includegraphics[width=0.22\linewidth]{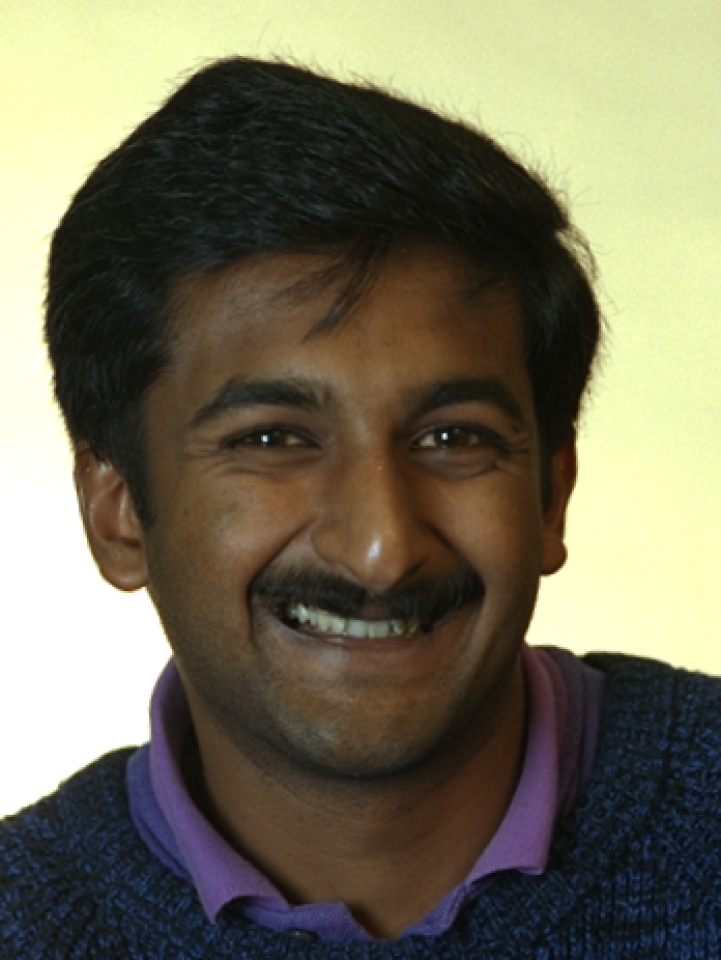}		
		\includegraphics[width=0.22\linewidth]{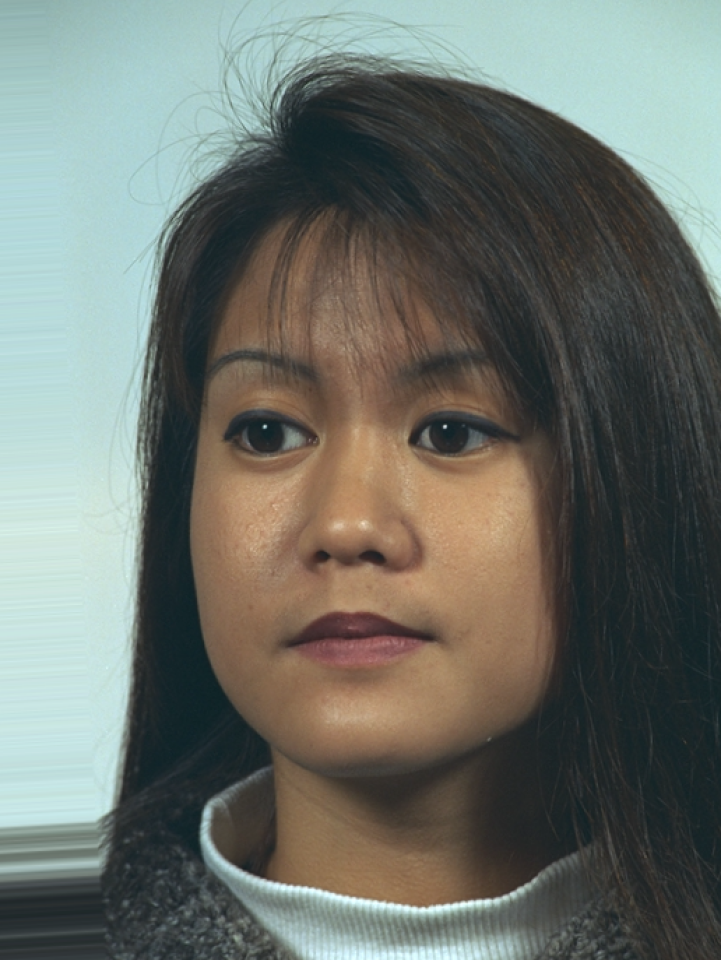}		
	}
	\subfigure[FRGCv2 probes]{
		\includegraphics[width=0.22\linewidth]{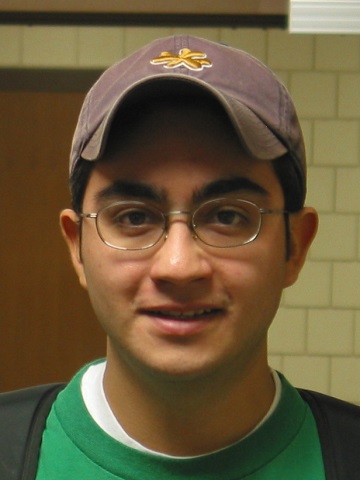}		
		\includegraphics[width=0.22\linewidth]{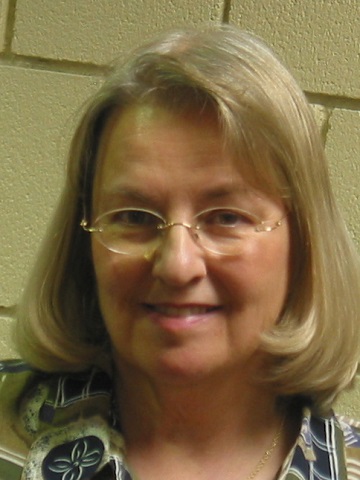}	
	}
	\caption{Examples of reference and  probe images of the used databases.}%
	\label{fig:example_dbs}\vspace{-0.2cm}%
\end{figure}

Experiments are conducted in a cross-database setting using manually selected subsets of the publicly available FERET \cite{Phillips1998} and FRGCv2 \cite{Phillips2005} databases. That is, the training process in which feature distributions are estimated to determine the quantization parameters, is conducted on the FERET database and the evaluation is performed on the FRGCv2 database and vice versa. Reference and probe images were chosen with the aim of simulating a cooperative authentication scenario where the enrolment process has been performed in a controlled environment. Precisely, reference images of both databases largely fulfill the requirements defined by International Civil Aviation Organization (ICAO) in \cite{InternationalCivilAviationOrganisation2006},  in particular frontal pose and neutral expression. In contrast, probe images exhibit variations in pose, expression, focus and illumination. If possible, probe images were preferably chosen from different acquisition sessions in order to obtain a realistic scenario. Examples of probe and reference images of both face image subsets are depicted in figure~\ref{fig:example_dbs}. Generally, the FRGCv2 subset contains less constrained images and is considered as more challenging, compared to the FERET subset as well as databases used in previously published works, see table~\ref{tab:related}. Table~\ref{tab:dbs} lists the number of subjects, corresponding reference and probe images, as well as the resulting mated and non-mated comparisons.

While this work considers a cooperative authentication scenario, the biometric performance and security of the proposed face fuzzy vault system could be evaluated on less-constrained databases, \eg Labelled Faces in the Wild (LFW) \cite{Learned-Miller16a}. While this is beyond the scope of this work, it is expected that a degradation of verification performance in an unprotected system also implies degradation of verification performance of the corresponding cryptosystem. This is generally confirmed in published works on biometric template protection where increased biometric performance error rates (FNMR and FMR) in the unprotected systems have been found to be also reflected in the protected system \cite{Rathgeb11e}. Furthermore, the application of template protection is usually reported to lead to inferior biometric performance compared to the original unprotected systems \cite{BNandakumar15a}. In other words, privacy protection often comes at the cost of recognition accuracy. 

\begin{table}[!t]
\centering
\caption{Overview of face image subsets from the FERET and FRGCv2 face databases:  amount of subjects, corresponding reference and probe images as well as resulting number of mated and non-mated comparisons  (``f'' and ``m'' denote female and male, respectively). }\label{tab:dbs}\vspace{-0.2cm}
\resizebox{.5\textwidth}{!}{
\renewcommand*{\arraystretch}{1.2}
\begin{tabular}{|l|l|c|c|c|c|}
\hline
\multirow{2}{*}{\textbf{Database}} & \multirow{2}{*}{\textbf{Subjects (f/m)}} & \multicolumn{2}{c|}{\textbf{Images}}  & \multicolumn{2}{c|}{\textbf{Comparisons}}  \\
 &  & \textbf{Reference}  & \textbf{Probe} & \textbf{Mated} &  \textbf{Non-Mated} \\\hline
FERET & 529 (200/329) &  529 & 791 & 791 & 139,128 \\\hline
FRGCv2 & 533 (231/302) & 984 & 1,726 & 3,298 & 141,246 \\\hline
\end{tabular}
}
\end{table}

\subsection{Binarisation Methods}
\label{sec:binmeth}

In the first experiment, combinations of quantisation and binarisation methods described in section~\ref{sec:featuretransform} are benchmarked against the baseline system. Feature elements of deep face representations extracted from images of the FERET and FRGCv2 databases were found to lie in the range $[-0.3, 0.3]$, which was used as basis for estimating equally-sized intervals. 
In the baseline system, comparison scores between pairs of deep face representations are obtained by estimating their Euclidean distance. Biometric performance is evaluated in terms of False Non-Match Rate (FNMR)  and False Match Rate (FMR) \cite{ISO-PerformanceReporting-2006}. In particular the Equal Error Rate (EER), \textit{i.e.} the operation point where \mbox{FNMR $=$ FMR}, is reported. Further, a measure of decidability ($d'$)~\cite{Daugman-DecisionLandscapes-TR-2000} calculated as 
\vspace{-0.1cm}
\begin{equation}
d'=\frac{|\mu_{g} - \mu_{i}|}{\sqrt{\frac{1}{2}(\sigma_{g}^{2}+\sigma_{i}^{2}})}
\end{equation}
 is reported, where $\mu_{g}$ and $\mu_{i}$ represent the means of the  mated and the non-mated comparison trials score distributions and $\sigma_{g}$ and $\sigma_{i}$ their standard deviations, respectively. Larger decidability values indicate better separability of mated and non-mated comparison scores. The aforementioned metrics are estimated using the PyEER python package\footnote{\url{https://pypi.org/project/pyeer/}} for biometric systems performance evaluation.

\begin{figure}[!t]
\vspace{-0.0cm}
\centering
\includegraphics[width=0.47\textwidth]{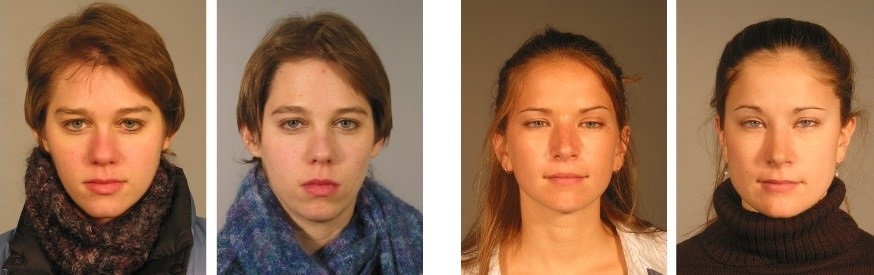}
\caption{Examples of pairs of images causing false matches in the FRGCv2 database.}\label{fig:twins}\vspace{-0.2cm}
\end{figure}

Table~\ref{tab:resultsbin} gives an overview of the obtained results. It can be observed that the baseline system achieves a perfect separation between mated and non-mated scores on the FERET database. On the FRGCv2 database, an extremely low EER value is obtained. When analyzing corresponding false matches it was found that those result from a handful image pairs of subjects which appear to be related or even monozygotic (identical) twins, see figure~\ref{fig:twins}. That is, reported EER values on the FRGCv2 database result from the interpolation of a few data points and, hence, shall be treated with caution. On both databases high decidability measures are obtained which are generally lower for the FRGCv2 database (which confirms that this database is more challenging). Further, it can be seen that the Boolean binarisation method reduces the discriminativity of deep face representations, in particular on the  FRGCv2 database. Neither the DBR nor the BRGC binarisation schemes improve upon the Boolean method. When using more intervals the performance generally decreases for these binarisation schemes.

\begin{table}[!t]
\centering
\caption{Performance results in terms of EER (in\%) $|$ $d'$ for different binarisation methods on both databases.}\label{tab:resultsbin}\vspace{-0.2cm}
\resizebox{.48\textwidth}{!}{
\renewcommand*{\arraystretch}{1.2}
\begin{tabular}{|c|c|r|r|r|r|}
\hline
\multirow{3}{*}{\textbf{Method}} & \multirow{2}{*}{\textbf{Intervals}}& \multicolumn{2}{c|}{\textbf{Training:} FRGCv2 } & \multicolumn{2}{c|}{\textbf{Training:} FERET}  \\
 & \multirow{2}{*}{$d$} & \multicolumn{2}{c|}{\textbf{Evaluation:} FERET} & \multicolumn{2}{c|}{\textbf{Evaluation:} FRGCv2}   \\
 &  & \multicolumn{1}{c|}{\textbf{equal-prob.}}  & \multicolumn{1}{c|}{\textbf{equal-size}} & \multicolumn{1}{c|}{\textbf{equal-prob.}} &  \multicolumn{1}{c|}{\textbf{equal-size}}   \\\hline
\multirow{1}{*}{Baseline} & -- &  \multicolumn{2}{c|}{0.0  $\,|\,$ 9.9}  &  \multicolumn{2}{c|}{0.001  $\,|\,$ 9.4}  \\\hline
\multirow{1}{*}{Boolean} & 2 & 0.0  $\,|\,$ 9.7 & 0.0  $\,|\,$ 9.5 &  0.003  $\,|\,$ 8.2 & 0.003  $\,|\,$ 8.2  \\\hline
\multirow{3}{*}{DBR} & 4 & 0.0  $\,|\,$ 7.2 & 0.0  $\,|\,$ 8.9  & 0.003  $\,|\,$ 7.5 &  0.003  $\,|\,$ 7.6   \\
 & 8 & 0.0  $\,|\,$ 6.6 & 0.0  $\,|\,$ 6.6 & 0.02  $\,|\,$ 7.1 &  0.03  $\,|\,$ 6.3  \\
 & 16 & 0.0  $\,|\,$ 6.4 & 0.001  $\,|\,$ 5.6  & 0.03  $\,|\,$ 6.7 & 0.03  $\,|\,$ 6.0    \\
\hline
\multirow{3}{*}{BRGC} & 4 & 0.0  $\,|\,$ 8.1 & 0.0  $\,|\,$ 9.2 &  0.002  $\,|\,$ 7.4 & 0.002  $\,|\,$ 8.0     \\
 & 8 & 0.0  $\,|\,$ 6.3  & 0.0  $\,|\,$ 8.8 &  0.01  $\,|\,$ 6.6 & 0.07  $\,|\,$ 8.0    \\
 & 16 & 0.0  $\,|\,$ 5.8 & 0.0  $\,|\,$ 6.4 &  0.003  $\,|\,$ 6.2 & 0.04  $\,|\,$ 6.5   \\
\hline
\multirow{3}{*}{LSSC} & 4 & \textbf{0.0  $\,|\,$ 10.2} & 0.0  $\,|\,$ 9.2 & \textbf{0.001  $\,|\,$ 9.1} &  0.002  $\,|\,$ 8.0  \\
 & 8 & \textbf{0.0  $\,|\,$ 10.3} & 0.0  $\,|\,$ 9.6 &  \textbf{0.07  $\,|\,$ 9.3} &  0.07  $\,|\,$ 8.8   \\
 & 16 & \textbf{0.0  $\,|\,$ 10.3} & 0.0  $\,|\,$ 9.8 &  \textbf{0.07  $\,|\,$ 9.3}  & 0.07  $\,|\,$ 9.1   \\
\hline
\multirow{3}{*}{One-hot} & 4 & 0.0  $\,|\,$ 7.0 & 0.0  $\,|\,$ 8.7  & 0.001  $\,|\,$ 7.1 & 0.002  $\,|\,$ 7.6    \\
 & 8 & 0.0  $\,|\,$ 5.3 & 0.0  $\,|\,$ 6.0 &  0.006  $\,|\,$ 5.9 &  0.01  $\,|\,$ 5.8  \\
 & 16 & 0.0  $\,|\,$ 4.6 & 0.11  $\,|\,$ 4.0 &  0.18  $\,|\,$ 5.1 &  0.19  $\,|\,$ 4.6  \\
\hline   
 \end{tabular}
}
\end{table}

In contrast to the aforementioned binarisation schemes, the biometric performance achieved by the LSSC method is close to that of the baseline system. Best results are obtained when dividing the feature space into equally probable intervals. Here, the biometric performance generally improves with the number of intervals into which the feature space is divided. However, biometric performance quickly converges such that a no significant improvements are observable when using more than 8 intervals. Note that in some cases this method even shows a better decidability than the baseline system. 

\begin{table*}[!t]
\centering
\caption{Performance results in terms of FNMR $|$ FMR (in \%) for the fuzzy vault scheme using Lagrange (LG), Reed-Solomon (RS), and Guruswami-Sudan (GS) decoding strategies on both databases.}\label{tab:fv_results}\vspace{-0.1cm}
\resizebox{.7\textwidth}{!}{
\renewcommand*{\arraystretch}{1.2}
\begin{tabular}{|c|c|c|c|c|c|c|c|}
\hline
\textbf{Intervals} & \textbf{Degree} &  \multicolumn{3}{c|}{\textbf{Training:} FRGCv2, \textbf{Test:} FERET} & \multicolumn{3}{c|}{\textbf{Training:} FERET, \textbf{Test:} FRGCv2}\\
$d$ & $k$ &\textbf{LG} & \textbf{RS} & \textbf{GS} & \textbf{LG} & \textbf{RS} & \textbf{GS}  \\\hline 
\multirow{14}{*}{4} & 16 & 0.0 $\,|\,$ 99.05 & 0.0 $\,|\,$ 99.70 & 0.0 $\,|\,$ 100.0 & 0.0 $\,|\,$ 99.23 & 0.0 $\,|\,$ 99.79 & 0.0 $\,|\,$ 100.0 \\
 &32 & 0.002 $\,|\,$ 0.33 & 0.0 $\,|\,$ 99.05 & 0.0 $\,|\,$ 100.0 & 0.30 $\,|\,$ 0.48 & 0.0 $\,|\,$ 99.29 & 0.0 $\,|\,$ 100.0\\
 & 48 & 3.26 $\,|\,$ 0.0 & 0.0 $\,|\,$ 97.48 & 0.0 $\,|\,$ 100.0 &  33.54 $\,|\,$ 0.001 & 0.0 $\,|\,$ 97.95 & 0.0 $\,|\,$ 100.0\\
 & 64 & 24.04 $\,|\,$ 0.0 & 0.0 $\,|\,$ 94.11 & 0.0 $\,|\,$ 100.0 & 85.86 $\,|\,$ 0.0 & 0.0 $\,|\,$ 95.05 & 0.0 $\,|\,$ 100.0\\
 & 96  & 72.99 $\,|\,$ 0.0 & 0.0 $\,|\,$ 77.41 & 0.0 $\,|\,$ 100.0 & 99.57 $\,|\,$ 0.0 & 0.0 $\,|\,$ 80.58 & 0.0 $\,|\,$ 100.0\\
  & 256 & 99.99 $\,|\,$ 0.0 & 0.0 $\,|\,$ 0.06 & 0.0 $\,|\,$ 58.94 &  99.99 $\,|\,$ 0.0 & 0.0 $\,|\,$ 0.28 & 0.0 $\,|\,$ 63.50\\
   & 288 & 99.99 $\,|\,$ 0.0 & 0.0 $\,|\,$ 0.004 & 0.0 $\,|\,$ 13.75  & 100.0 $\,|\,$ 0.0 & 0.0 $\,|\,$ 0.035 & 0.0 $\,|\,$ 18.36\\
  &320 & 100.0 $\,|\,$ 0.0 & 0.0 $\,|\,$ 0.001 & 0.0 $\,|\,$ 1.033  & 100.0 $\,|\,$ 0.0 & 0.061 $\,|\,$ 0.006 & 0.0 $\,|\,$ 2.277 \\
   & 352 &  100.0 $\,|\,$ 0.0 & 0.0 $\,|\,$ 0.0 & 0.0 $\,|\,$ 0.031 & 100.0 $\,|\,$ 0.0 & 0.819 $\,|\,$ 0.001 & 0.0 $\,|\,$ 0.154\\
  & 384 & 100.0 $\,|\,$ 0.0 & 0.0 $\,|\,$ 0.0 & 0.0 $\,|\,$ 0.001 & 100.0 $\,|\,$ 0.0 & 4.639 $\,|\,$ 0.001 & 0.061 $\,|\,$ 0.009\\
  & 416 & 100.0 $\,|\,$ 0.0 & 1.263 $\,|\,$ 0.0 & 0.0 $\,|\,$ 0.0&   100.0 $\,|\,$ 0.0 & 13.06 $\,|\,$ 0.001 & 1.637 $\,|\,$ 0.001\\
   & 448 & 100.0 $\,|\,$ 0.0 & 2.273 $\,|\,$ 0.0 & 0.758 $\,|\,$ 0.0  & 100.0 $\,|\,$ 0.0 & 33.08 $\,|\,$ 0.0 & 10.21 $\,|\,$ 0.001\\
  &480 & 100.0 $\,|\,$ 0.0 & 7.449 $\,|\,$ 0.0 & 2.146 $\,|\,$ 0.0 & 100.0 $\,|\,$ 0.0 & 59.15 $\,|\,$ 0.0 & 31.71 $\,|\,$ 0.0\\
   &512 & 100.0 $\,|\,$ 0.0 & 16.41 $\,|\,$ 0.0 & 8.586 $\,|\,$ 0.0 & 100.0 $\,|\,$ 0.0 & 82.01 $\,|\,$ 0.0 & 62.85 $\,|\,$ 0.0\\
\hline
\multirow{14}{*}{8} & 16 & 0.0 $\,|\,$ 99.99 & 0.0 $\,|\,$ 100.0 & 0.0 $\,|\,$ 100.0 & 0.0 $\,|\,$ 99.99 & 0.0 $\,|\,$ 100.0 & 0.0 $\,|\,$ 100.0 \\
 &32 & 0.0 $\,|\,$ 2.750 & 0.0 $\,|\,$ 100.0 & 0.0 $\,|\,$ 100.0 & 0.008 $\,|\,$ 3.520 & 0.0 $\,|\,$ 100.0 & 0.0 $\,|\,$ 100.0\\
 & 48 & 0.996 $\,|\,$ 0.003 & 0.0 $\,|\,$ 100.0 & 0.0 $\,|\,$ 100.0 &  13.41 $\,|\,$ 0.005 & 0.0 $\,|\,$ 100.0 & 0.0 $\,|\,$ 100.0\\
 & 64 & 11.26 $\,|\,$ 0.0 & 0.0 $\,|\,$ 100.0 & 0.0 $\,|\,$ 100.0 & 68.41 $\,|\,$ 0.0 & 0.0 $\,|\,$ 100.0 & 0.0 $\,|\,$ 100.0\\
 & 96 & 60.45 $\,|\,$ 0.0 & 0.0 $\,|\,$ 99.99 & 0.0 $\,|\,$ 100.0 & 98.72 $\,|\,$ 0.0 & 0.0 $\,|\,$ 100.0 & 0.0 $\,|\,$ 100.0\\
 & 512 & 100.0 $\,|\,$ 0.0 & 0.0 $\,|\,$ 18.84 & 0.0 $\,|\,$ 99.99 &  100.0 $\,|\,$ 0.0 & 0.0 $\,|\,$ 24.23 & 0.0 $\,|\,$ 99.99\\
 & 576 & 100.0 $\,|\,$ 0.0 & 0.0 $\,|\,$ 4.657 & 0.0 $\,|\,$ 99.37 & 100.0 $\,|\,$ 0.0 & 0.0 $\,|\,$ 7.728 & 0.0 $\,|\,$ 99.52\\
 & 640 & 100.0 $\,|\,$ 0.0 & 0.0 $\,|\,$ 0.723 & 0.0 $\,|\,$ 87.63 &  100.0 $\,|\,$ 0.0 & 0.0 $\,|\,$ 1.774 & 0.0 $\,|\,$ 89.68\\
  & 704 & 100.0 $\,|\,$ 0.0 & 0.0 $\,|\,$ 0.070 & 0.0 $\,|\,$ 44.72  & 100.0 $\,|\,$ 0.0 & 0.0 $\,|\,$ 0.283 & 0.0 $\,|\,$ 50.51\\
  & 768 & 100.0 $\,|\,$ 0.0 & 0.0 $\,|\,$ 0.004 & 0.0 $\,|\,$ 9.033 &  100.0 $\,|\,$ 0.0 & 0.0 $\,|\,$ 0.041 & 0.0 $\,|\,$ 13.08\\
  & 832 & 100.0 $\,|\,$ 0.0 & 0.0 $\,|\,$ 0.001 & 0.0 $\,|\,$ 0.697  & 100.0 $\,|\,$ 0.0 & 0.030 $\,|\,$ 0.006 & 0.0 $\,|\,$ 1.666\\
  & 896 & 100.0 $\,|\,$ 0.0 & 0.0 $\,|\,$ 0.0 & 0.0 $\,|\,$ 0.019  & 100.0 $\,|\,$ 0.0 & 0.424 $\,|\,$ 0.002 & 0.0 $\,|\,$ 0.130 \\
  & 960 & 100.0 $\,|\,$ 0.0 & 0.126 $\,|\,$ 0.0 & 0.0 $\,|\,$ 0.001 & 100.0 $\,|\,$ 0.0 & 2.941 $\,|\,$ 0.001 & 0.061 $\,|\,$ 0.009\\
  & 1024 & 100.0 $\,|\,$ 0.0 & 0.631 $\,|\,$ 0.0 & 0.0 $\,|\,$ 0.0 & 100.0 $\,|\,$ 0.0 & 9.096 $\,|\,$ 0.001 & 1.092 $\,|\,$ 0.001\\
\hline

 \end{tabular}
}
\end{table*}

The one-hot method obtains the worst biometric performance. Similar to the DBR and BRGC schemes, for this binarisation approach performance rates decrease if a larger number of intervals is used.

Based on this first experiment it can be concluded that the LSSC-based binarisation scheme with equally probable intervals is the most suitable method as it maintains the biometric performance of the baseline system. This confirms the findings in \cite{Drozdowski-DeepFaceBinarisation-ICIP-2018}, where similar results were reported for other DCNN-based face representations. Therefore, only this binarisation approach with up to 8 intervals will be considered for the construction of the fuzzy vault scheme in the subsequent experiments. 
 \begin{figure}[!t]
\vspace{-0.0cm}
\hspace{0.75cm}\includegraphics[width=7cm]{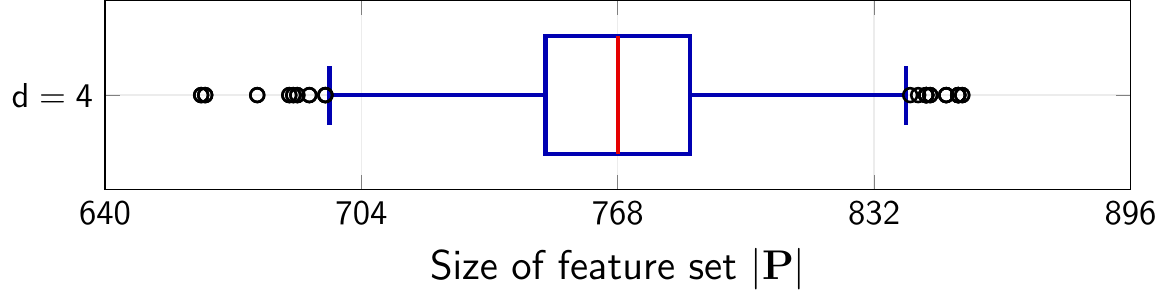}
\vspace{-0.2cm}
\hspace{0.75cm}\includegraphics[width=7.1cm]{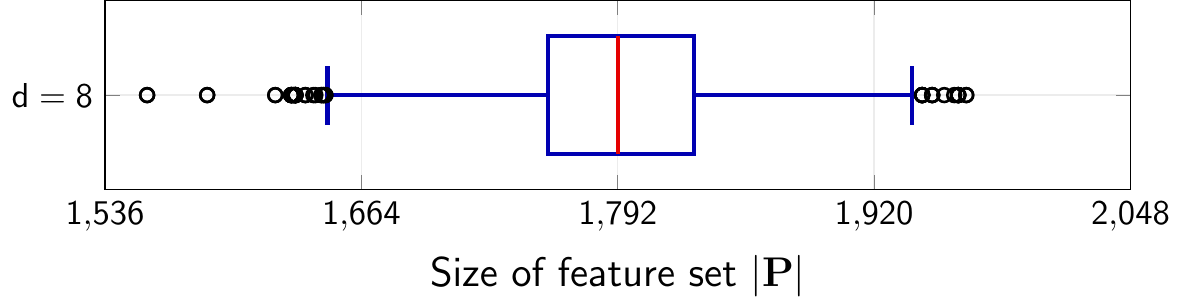}
\caption{Box plots of feature set sizes across both databases. }\label{fig:set_sizes}\vspace{-0.2cm}
\end{figure}

\subsection{Fuzzy Vault Construction and Decoding Strategy}
\label{sec:fcsdec}

Firstly, the sizes of feature sets are estimated. For equal probable intervals the expected size of a feature set is $E(|\mathbf{P}|)=|\mathbf{b}|/2$ with $|\mathbf{b}|=n(d-1)$ for the LSSC method  and $n=512$ for the used ArcFace system. Figure \ref{fig:set_sizes} depicts box plots of distributions of feature set sizes for different numbers of intervals across both used databases. It can be observed that set sizes are narrowly-distributed around their expected values with only a few mild outliers. In fact, feature set sizes approximate a binomial distribution $B(n(d-1);0.5)$, \ie
the probability of feature set sizes which are considerably smaller or larger than the expected value quickly diminishes. This narrow distribution of feature set sizes implies that the maximum observed vault size will be significantly below the theoretical maximum which in turn reduces storage requirement.  From figure \ref{fig:set_sizes}, it can be observed that set sizes quickly increase with the number of used intervals, \cf figure~\ref{fig:tempate_sizes}.

Table \ref{tab:fv_results} gives an overview of obtained biometric performance in terms of FNMR and FMR for the fuzzy vault scheme employing different decoding strategies on both databases. For the Lagrange decoder $2^{16}$ decoding attempts are performed where best performance rates are obtained at rather small polynomial degrees of approximately 32 and 48 for using 4 and 8 quantisation intervals, respectively. The Reed-Solomon decoder achieves competitive performance rates at  higher polynomial degrees, \ie approximately 320 for 4 intervals and 900 for 8 intervals. Comparable performance rates are obtained for applying the Guruswami-Sudan decoding with a multiplicity of 1, although at even higher polynomial degrees. Specifically, on the FERET database, a perfect separation between mated and non-mated decoding attempts is maintained, \eg for $d=4$ and $k=416$. For $d=4$ and $k=384$, a FNMR of 0.06\% at a FMR$<$0.01\% is achieved on the more challenging FRGCv2 database. In summary, it can be observed that the use of a Reed-Solomon and a Guruswami-Sudan decoding strategies yield highest recognition accuracies.

\subsection{Security and Runtime Analysis}
\label{sec:runtime}

Security is measured in terms of False Accept Security (FAS). As mentioned earlier, compared to the FAS, the Brute-Force Security (BFS) tends to significantly overestimate the effective security of a fuzzy vault scheme \cite{bib:TamsMihailescuMunk2015}. The BFS usually increases with the degree of the secret polynomial. Precisely, for the proposed system the minimum BFS observed (in bits) was similar to the size of the polynomial, \ie $\min(\mbox{BFS}) \approx k$. Therefore, a more realistic measure can be derived from the FMR. An attacker can iteratively simulate non-mated verification attempts until the vault is unlocked thereby running a false-accept attack; within each simulated attempt, the probability of success equals the FMR. The FAS is estimated as,

\vspace{-0.4cm}
\begin{equation}
l \cdot \log(0.5)/\log(1-\mbox{FMR})
\end{equation}
where $l$ is the average amount of operations for a non-mated verification attempt.  Precisely, the FAS defines the number of operations that an attacker requires to succeed with a probability of 50\% (alternatively, the FAS could be estimated as $l/$FMR, \ie the expected number of steps until an attack succeeds. Note that $l$ depends on the chosen $k$ and is measured in terms of Lagrange interpolation. That is, for the remaining decoding strategies $l$ is measured relatively to the Lagrange interpolation.
From table~\ref{tab:fv_results}, it can be observed that, for any polynomial degree, the Guruswami-Sudan decoding strategy yields the highest FMR. Therefore, it is reasonable to assume that an attacker would use the Guruswami-Sudan algorithm during a false-accept attack. The amount of operations needed for the Guruswami-Sudan decoding algorithm is measured relatively to the Lagrange method. Precisely, we express the effort of an attack by the number of Lagrange interpolations that would (roughly) result in the same computational time. For different polynomial degrees the computational times required by the used decoding strategies was empirically estimated. Note that this results in a rather conservative measure, since a single Lagrange interpolation for the used degrees $k$ is expected to be more time consuming than a decryption attempt within a classical cryptographic method, \eg AES.

However, for some polynomial degrees no false matches have been observed, see table~\ref{tab:fv_results}. In such cases, the FAS can not be estimated and is approximated by linearly interpolating the FASs of the last two polynomial degrees for which the FAS could be estimated. Alternatively, other approximations, \eg rule of three \cite{ISO-IEC-19795-1:2006}, could be applied. But the rule of three for example, does not consider the fact that security is expected to increase as $k$ increases. Table \ref{tab:fas_results} summarises FASs in relation to the Genuine Match Rate (GMR$=$1$-$FNMR) provided by the fuzzy vault scheme using the Guruswami-Sudan decoding strategy (interpolated FAS values are marked italic). On the FERET database, a FASs of approximately 25 bits and 30 bits are obtained for a GMRs of 100\% and 95\%, respectively, in case 4 intervals are used. For the same GMRs, on the more challenging FRGCv2 database,  FASs of around 20 bits and 22 bits are achieved. For the use of 8 intervals, slightly higher FASs are approximated on both databases.

\begin{table}[!t]
\centering
\caption{Performance in relation to security in terms of GMR (in \%) $|$ FAS (in bits) and decoding time in terms of $t_g$ $|$ $t_i$  (in ms) for the fuzzy vault scheme using the Guruswami-Sudan decoding strategy on both databases.}\label{tab:fas_results}\vspace{-0.1cm}
\resizebox{.49\textwidth}{!}{
\renewcommand*{\arraystretch}{1.2}
\begin{tabular}{|c|c|c|c|c|c|c|c|c|}
\hline
\textbf{Intervals} & \textbf{Degree} &  \multicolumn{1}{c|}{\textbf{Training:} FRGCv2, } & \multicolumn{1}{c|}{\textbf{Training:} FERET, } & \multirow{2}{*}{\textbf{Decoding time}}\\
$d$ & $k$ & \textbf{Test:} FERET&  \textbf{Test:} FRGCv2  & \\\hline 
\multirow{13}{*}{4}  & 320 & 100.00 $\,|\,$ 12.1 & 100.00 $\,|\,$ 11.0 & 82.5 $\,|\,$ 68.2  \\
& 336 & 100.00 $\,|\,$ 14.5 & 100.00 $\,|\,$ 12.8 & 80.1 $\,|\,$ 68.1  \\
& 352 & 100.00 $\,|\,$ 17.2 & 100.00 $\,|\,$ 14.9 & 80.2 $\,|\,$ 68.4  \\
& 368 & 100.00 $\,|\,$ 20.3 & 99.97 $\,|\,$ 16.9 & 80.8 $\,|\,$ 68.8  \\
& 384 & 100.00 $\,|\,$ 21.7 & 99.94 $\,|\,$ 19.0 & 81.9 $\,|\,$ 70.1  \\
& 400 & 100.00 $\,|\,$ \textit{23.0} & 99.52 $\,|\,$ 19.9 & 82.7 $\,|\,$ 70.5  \\
& 416 & 100.00 $\,|\,$ \textit{24.4} & 98.36 $\,|\,$ 21.7 & 83.4 $\,|\,$ 69.1  \\
& 432 & 99.75 $\,|\,$ \textit{25.7} & 95.18 $\,|\,$ 21.7 & 83.9 $\,|\,$ 69.2  \\
& 448 & 99.24 $\,|\,$ \textit{27.1} & 89.78 $\,|\,$ 21.7 & 84.7 $\,|\,$ 70.9  \\
& 464 & 98.61 $\,|\,$ \textit{28.4} & 81.57 $\,|\,$ 22.7 & 85.1 $\,|\,$ 70.3  \\
& 480 & 97.85 $\,|\,$ \textit{29.8} & 68.28 $\,|\,$ \textit{23.7} & 83.6 $\,|\,$ 71.2  \\
& 496 & 95.33 $\,|\,$ \textit{31.1} & 52.64 $\,|\,$ \textit{24.7} & 83.3 $\,|\,$ 70.4  \\
& 512 & 91.41 $\,|\,$ \textit{32.5} & 37.14 $\,|\,$ \textit{25.6} & 83.0 $\,|\,$ 70.2  \\
\hline
\multirow{13}{*}{8} & 1008 & 100.00 $\,|\,$ \textit{26.9} & 99.49 $\,|\,$ 23.3 & 401.0 $\,|\,$ 359.3  \\
& 1024 & 100.00 $\,|\,$ \textit{27.9} & 98.91 $\,|\,$ 24.4 & 402.3 $\,|\,$ 360.2  \\
& 1040 & 99.87 $\,|\,$ \textit{28.9} & 98.12 $\,|\,$ 24.5 & 403.6 $\,|\,$ 364.1  \\
& 1056 & 99.87 $\,|\,$ \textit{29.8} & 96.60 $\,|\,$ 24.5 & 406.0 $\,|\,$ 363.5  \\
& 1072 & 99.62 $\,|\,$ \textit{30.8} & 95.00 $\,|\,$ 24.6 & 413.0 $\,|\,$ 364.5  \\
& 1088 & 99.62 $\,|\,$ \textit{31.8} & 92.94 $\,|\,$ 24.6 & 419.6 $\,|\,$ 365.5  \\
& 1104 & 99.50 $\,|\,$ \textit{32.8} & 90.27 $\,|\,$ 24.7 & 417.2 $\,|\,$ 374.0  \\
& 1120 & 98.86 $\,|\,$ \textit{33.7} & 86.66 $\,|\,$ 25.7 & 419.5 $\,|\,$ 367.4  \\
& 1136 & 98.61 $\,|\,$ \textit{34.7} & 82.29 $\,|\,$ 25.8 & 423.4 $\,|\,$ 369.3  \\
& 1152 & 98.23 $\,|\,$ \textit{35.7} & 76.90 $\,|\,$ 25.8 & 426.3 $\,|\,$ 376.1  \\
& 1168 & 97.85 $\,|\,$ \textit{36.7} & 69.89 $\,|\,$ \textit{25.9} & 429.6 $\,|\,$ 373.9  \\
& 1184 & 97.10 $\,|\,$ \textit{37.6} & 62.64 $\,|\,$ \textit{25.9} & 432.0 $\,|\,$ 378.3  \\
& 1200 & 95.83 $\,|\,$ \textit{38.6} & 55.25 $\,|\,$ \textit{25.9} & 442.9 $\,|\,$ 397.1  \\\hline
 \end{tabular}
}
\end{table}

\begin{figure}[!t]
\centering
\includegraphics[width=0.425\textwidth]{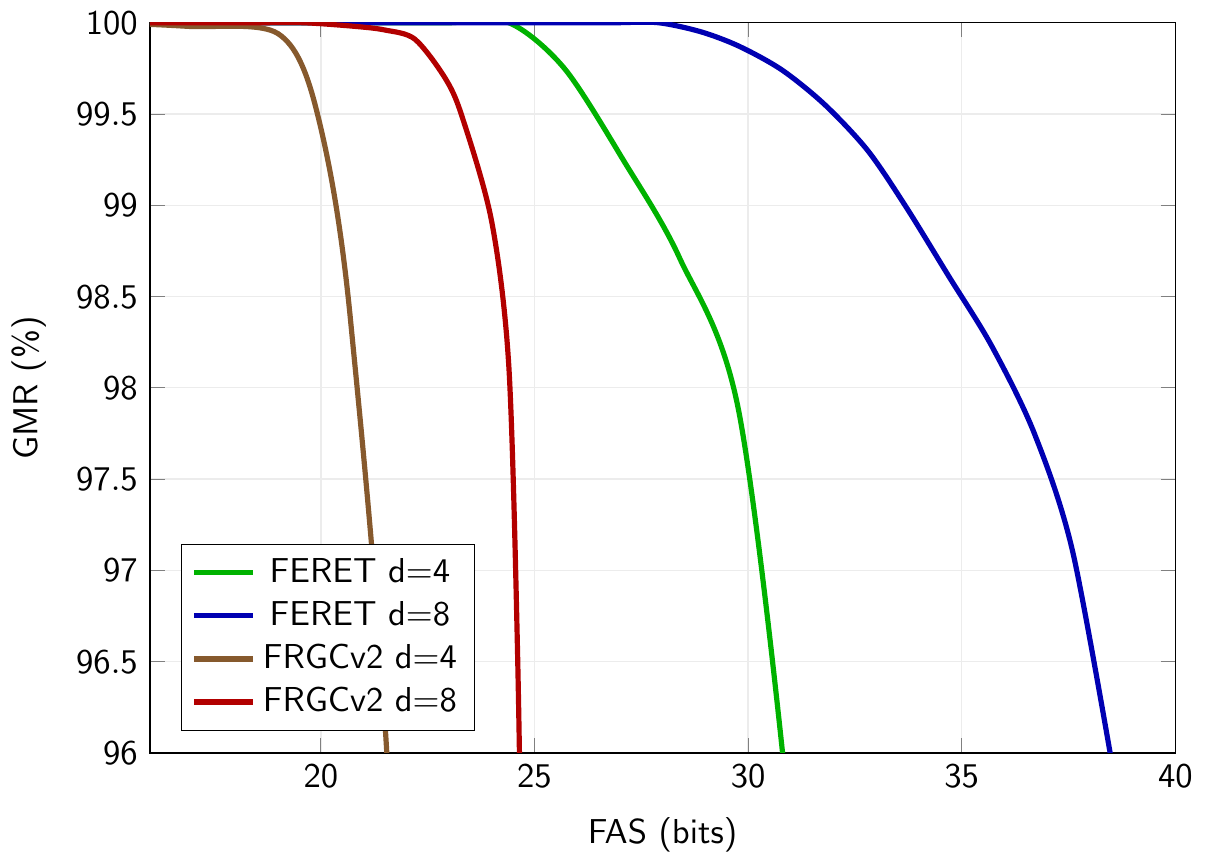}\vspace{-0.2cm}
\caption{Performance rates in relation to the FAS.}\label{fig:gmr_far}\vspace{-0.4cm}
\end{figure}

Figure~\ref{fig:gmr_far} illustrates the relation between GMR and FAS on both databases. It can be seen that the employed approximation results in steeply descending curves, thus providing a conservative approximation of the FAS for polynomial degrees where no false match is observed. At certain polynomial degrees, only a handful false matches occur. As mentioned earlier these false matches likely result from image pairs of subjects which appear to be siblings or even monozygotic twins. It is important to note that this is a well-documented limitation in face recognition which also represents a security threat to the proposed system.

\setlength{\tabcolsep}{2.5pt}
\begin{table*}[!t]
\centering
\caption{Comparison with most relevant works on face-based fuzzy vault schemes.}\label{tab:related}\vspace{-0.1cm}
\resizebox{\textwidth}{!}{
\renewcommand*{\arraystretch}{1.2}
\begin{tabular}{|c|c|c|c|c|c|c|}
\hline
\textbf{Approach} & \textbf{Database} & \textbf{Feature Extraction} & \textbf{Feature Transformation} & \textbf{Decoding Strategy} & \textbf{Performance Rates} & \textbf{Security Rates} \\\hline
\begin{tabular}{@{}c@{}}Feng \textit{et al.}\\ \cite{Feng06a} \end{tabular}&  \begin{tabular}{@{}c@{}} ORL face database\\ (40 subjects)\end{tabular} & \begin{tabular}{@{}c@{}}Eigenfaces and\\ Fisherfaces with LDA\end{tabular} & \begin{tabular}{@{}c@{}} Quantisation of feature\\ vector segments  \end{tabular} &  Reed Solomon &  \begin{tabular}{@{}c@{}}$\sim$5\% EER\\(non-stolen token\\ scenario) \end{tabular} & \begin{tabular}{@{}c@{}}41 bits brute force\\ security\end{tabular} \\\hline
\begin{tabular}{@{}c@{}}Wang  and\\ Plataniotis\\ \cite{Wang07}  \end{tabular}& \begin{tabular}{@{}c@{}} ORL face database \end{tabular} & \begin{tabular}{@{}c@{}}Eigenfaces with PCA\\  \end{tabular} & \begin{tabular}{@{}c@{}} Random transformation,\\ feature quantisation,\\ binarisation \end{tabular} & \begin{tabular}{@{}c@{}}Cyclic Redundancy\\ Check\end{tabular} &  \begin{tabular}{@{}c@{}} FNMR=0.5\%,\\ FMR=7.38\%\end{tabular}  &  \begin{tabular}{@{}c@{}}$\sim$ 57 bits brute force\\ security\end{tabular}  \\\hline 
\begin{tabular}{@{}c@{}}Frassen \textit{et al.}\\ \cite{Frassen08}  \end{tabular} &  \begin{tabular}{@{}c@{}} 3D face database \\(100 subjects) \end{tabular} & \begin{tabular}{@{}c@{}}Depth-histogram-based\\ feature extraction\end{tabular}  & \begin{tabular}{@{}c@{}} Binarisation, quantisation\\ of bit chunks with address\\ bits  \end{tabular} & Least Squares Fitting  & \begin{tabular}{@{}c@{}}FNMR=4.1\%,\\ FMR=0.0\%\end{tabular}   & \begin{tabular}{@{}c@{}}$\sim$ 90 bits brute force\\ security\end{tabular}   \\\hline
\begin{tabular}{@{}c@{}}Wu and\\ Yuan\\ \cite{Wu10a}  \end{tabular} & \begin{tabular}{@{}c@{}} ORL face database \end{tabular} &  Eigenfaces with PCA & \begin{tabular}{@{}c@{}}Random transformation,\\ quantisation of feature \\ elements \end{tabular}& \begin{tabular}{@{}c@{}}Cyclic Redundancy\\ Check\end{tabular} & \begin{tabular}{@{}c@{}} FNMR=21.0\%,\\ FMR=16.5\%\end{tabular} &  \begin{tabular}{@{}c@{}}n.a. \end{tabular}\\\hline
\begin{tabular}{@{}c@{}}Nagar \textit{et al.}\\ \cite{BNagar12a} \end{tabular} & \begin{tabular}{@{}c@{}} XM2VTS and WVU\\ databases\\ (100/138 subjects) \end{tabular} & \begin{tabular}{@{}c@{}}Histogram-equalisation\\ with LDA \end{tabular}  & \begin{tabular}{@{}c@{}}Unary encoding-based \\binarisation, quantisation\\ of bit chunks\end{tabular}  & \begin{tabular}{@{}c@{}}Berlekamp-Massey\\ for Reed-Solomon\end{tabular} &  \begin{tabular}{@{}c@{}}FNMR=33\%/42\%\\ (WVU/XM2VTS) \end{tabular} & \begin{tabular}{@{}c@{}}51 bits false accept\\ security \end{tabular}\\\hline
\begin{tabular}{@{}c@{}}Dong \textit{et al.}\\ \cite{Dong21} \end{tabular} & \begin{tabular}{@{}c@{}} LFW, VGG, and\\  IJB-C databases\\ (610/9,131/3,530 subjects) \end{tabular} & \begin{tabular}{@{}c@{}}DCNN\\ (FaceNet,ArcFace) \end{tabular}  & \begin{tabular}{@{}c@{}}Index-of-maximum \\hashing\end{tabular}  & \begin{tabular}{@{}c@{}}Lagrange\\ Interpolation\end{tabular} &  \begin{tabular}{@{}c@{}}R1=99.9\%/99.8\%/81.4\%\\ (LFW/VGG/IJB-C) \end{tabular} & \begin{tabular}{@{}c@{}}37 bits false accept\\ security \end{tabular}\\\hline
\begin{tabular}{@{}c@{}}This work \end{tabular} & \begin{tabular}{@{}c@{}} FERET and FRGCv2\\ databases (subsets)\\ (529/533 subjects) \end{tabular} & \begin{tabular}{@{}c@{}}DCNN \\(ArcFace) \end{tabular}  & \begin{tabular}{@{}c@{}}Quantisation, LSSC-based \\binarisation of feature\\ elements,  mapping of \\binary code to feature set\\  \end{tabular}  & \begin{tabular}{@{}c@{}}Guruswami-Sudan\end{tabular} &  \begin{tabular}{@{}c@{}}$<$1\% FNMR at \\$<$0.01\% FMR\\ (FERET/FRGCv2) \end{tabular} & \begin{tabular}{@{}c@{}} $\sim$ 32 bits (FERET)\\ $\sim$ 24 bits (FRGC)\\ false accept \\security \end{tabular}\\\hline
\end{tabular}\vspace{-0.2cm}
}
\end{table*}
Another threat are correlation attacks that combine two or more vaults derived from the same subject. The correlation attack of \cite{bib:ScheirerBoult2007,bib:KholmatovYanikoglu2008} on the original fuzzy vault scheme, which represents a special linkage attack, can not be applied to the improved fuzzy vault scheme. It has been shown in \cite{bib:MerkleTams2013} that two related vaults $V(X)$ and $W(X)$ protecting the feature sets $\mathbf{P}$ and $\mathbf{P}'$, respectively,  can be attacked efficiently and effectively based on the extended Euclidean algorithm, provided that 
\begin{equation}\label{eq:cmreq}
|\mathbf{P}\cap\mathbf{P}'|\geq(\max(|\mathbf{P}|,|\mathbf{P}'|)+k)/2 \mbox{.} 
\end{equation}
This attack is prevented by applying the bijection \cite{bib:MerkleTams2013}, \ie re-mapping of feature elements. Due to the assumed randomness of two bijections $\sigma$ and $\sigma'$, the corresponding sets $\sigma(\mathbf{P})$ and $\sigma'(\mathbf{P}')$ are random and, based on the definition of the hyper-geometric distribution, the probability that for these sets Eq. \ref{eq:cmreq} is fulfilled is equal to 
\begin{equation}\label{eq:new}
1-{\rho\choose|\mathbf{P}|}^{-1}\sum_{j=0}^{\omega_0-1}{|\mathbf{P}'|\choose j}{\rho-|\mathbf{P}'|\choose |\mathbf{P}|-j} \mbox{ ,}
\end{equation}
where $\omega_0=\lceil(|\mathbf{P}|+k)/2\rceil$, $\rho=d|\mathbf{P}|$, and w.l.o.g. $|\mathbf{P}|\geq|\mathbf{P}'|$. For the used feature extractor and the feature transformation this probability is negligible. Hence, the requirement of unlinkability is fulfilled.

In order to obscure the size of the vault, which might leak information about the protected deep face representation, it is suggested choose a random polynomial  $Q(X)$ of degree $nm-t$ which should not exhibit zeros in $\mathbf{F}$. Subsequently, $V(X)$ can be defined as $V(X)=\kappa(X) + Q(X) \cdot \prod_{v \in \hat{\mathbf{P}}} (X-v)$ and thereby, it will always be of degree $nm$.

Finally, note that a narrow distribution of feature set sizes provides a similar level of security across users where the minimum security is expected to be close to the average, \cf figure~\ref{fig:set_sizes}. Note that this is usually not guaranteed for other biometric feature sets, \eg minutiae sets.  

In addition, table~\ref{tab:fv_results} lists the average decoding times  for mated ($t_g$) and non-mated ($t_i$) decoding attempts (excluding the time required for feature extraction). Runtime measures were conducted on a single core of an Intel\textsuperscript{\scriptsize\textregistered} Core\textsuperscript{\scriptsize TM} i5-8250U CPU at 1.60 GHz. Efficient decoding times which are significantly below 100~ms are obtained for the use of 4 intervals and polynomial sizes up to 512. For 8 intervals and polynomial degrees of less than 1200, decoding times below 500~ms are achieved. That is, key retrieval can be performed in real time for relevant parameters of the proposed fuzzy vault scheme which is essential for the usability of the system.  

\subsection{Comparison with other Works}
\label{sec:comprelated}

Table~\ref{tab:related} provides a comparison of the most relevant works against the proposed system. It can be observed that the presented fuzzy vault scheme significantly outperforms  published approaches based on hand-crafted feature extractors in terms of biometric performance. Further, it is worth noting that in contrast to the listed works, the proposed scheme is evaluated on more challenging databases. Focusing on security in terms of BFS, the presented system would outperform most published approaches by orders of magnitude. Furthermore, the FMRs reported in \cite{Feng06a,Wang07,Wu10a} show that the FAS of these schemes is extremely low  which confirms our assertion that BFS is not a useful measure for the actual security, \ie against arbitrary attacks. 

The comparison with \cite{BNagar12a}  requires some context: the FAS of 51 bits reported therein are obtained for unpractical FNMRs above 30\% while at a GMR of 95\% the FAS decreases below 22 bits. Moreover, the security estimates in \cite{BNagar12a} assume that an attacker applies essentially the same decoding strategy (\ie a Reed-Solomon decoder) as used for verification attempts, which is an unrealistic assumption; an attacker deploying a different strategy (\eg using a Guruswami-Sudan decoder) could have a much lower workload resulting in a drastically reduced security estimate. Furthermore, the computational effort for a genuine decoding (verification) attempt in \cite{BNagar12a} is very high and increases with the security parameters; it is, thus, questionable whether this scheme is practical for higher security levels. That is, the proposed system is expected to outperforms existing face-based fuzzy vault schemes which is confirmed by the extrapolated FAS. Therefore, we conclude that the proposed system outperforms existing face-based fuzzy vault schemes in terms of recognition accuracy and security. 

Very recently, a so-called chaff-less fuzzy vault scheme which utilises DCNNs for feature extraction was proposed in \cite{Dong21}. In contrast to the proposed scheme, the system in   \cite{Dong21} was designed for facial identification (one-to-many search) and was evaluated in a closed-set scenario in terms of rank-1 (R1) recognition rate. This hampers a direct comparison with the presented approach. However, it is worth to note that a closed-set identification scenario is generally considered less challenging and, hence, less realistic as it does not consider non-mated identification trails. This may also imply that the 37 bits FAS reported in  \cite{Dong21} drops in an open-set identification scenario.  

The security level achieved by the proposed scheme is very high for a biometric cryptosystem based on a single biometric characteristic. Fuzzy vault schemes utilizing a single fingerprint, \eg in \cite{bib:NandakumarJainPankanti2007,bib:TamsMihailescuMunk2015}, have been reported to achieve security levels of up to approximately 20 bits at a GMR of 90\%. As shown in \cite{Rathgeb-ImprovedMultiBiometrics-EURASIP-2016}, for a single iris fuzzy vault a FAS of around 35 bits can be achieved for a GMR of 95\%. Recent works which apply the fuzzy vault scheme to other physiological biometric characteristics obtain similar recognition accuracy which suggests that they provide similar security levels, \eg a palmprint-based fuzzy vault scheme in \cite{LENG20152290}. In \cite{mci/Korte2008} a security level of 73 bits has been achieved by a fuzzy vault scheme based on genetic fingerprints (DNA), but this characteristic is not relevant for real-time applications. Of course, multi-biometric schemes can provide higher security levels. For instance, for the use of four fingerprints in \cite{bib:Tams2015} and two irises \cite{Rathgeb-ImprovedMultiBiometrics-EURASIP-2016}, a false accept security of  65  and 57 bits have been reported at a FNMR of 7\%, respectively. In contrast, fuzzy vault schemes based on behavioural biometric characteristics, \eg online signature \cite{8954723} or accelerometer-based biometric features  \cite{7934407}, usually report significantly lower biometric performance mainly due to high intra-class variance. This indicates that such schemes also provide lower security levels.

\subsection{Potential Improvements}
\label{sec:improvements}

The proposed system outperforms published works on face-based fuzzy vault schemes in terms of biometric performance, \cf table~\ref{tab:related}. However, in order to provide higher security in terms of FAS, the fuzzy vault scheme could be hardend using a password or a multi-biometric fuzzy vault scheme could be constructed using further biometric characteristics, as suggested in \cite{BNagar12a}. Since the proposed transformation method can be applied to any fixed-length real valued feature vectors, an extension of this work to a multi-biometric fuzzy vault using feature-level fusion with neural networks is straightforward.

The runtime of the employed polynomial reconstruction method might be further improved. \Eg in \cite{bib:Tams2015}, it is suggested to initially apply a classical Reed-Solomon decoder to recover the correct polynomial; if unsuccessful, a Guruswami-Sudan algorithm is iteratively applied with increasing multiplicity until the correct polynomial is found or a maximum multiplicity is reached. It was shown that thereby, decoding times for mated verification attempts can be significantly decreased. A similar strategy could be used in the proposed system.
	
\section{Conclusion}
\label{sec:conclusion}

This work presented an unlinkable improved fuzzy vault-based template protection scheme for deep feature representations. For this purpose, a biometric characteristic-agnostic  feature transformation method was proposed which transforms fixed-length real-valued feature vectors to integer-valued feature sets. To this end, a comprehensive analysis of various feature quantisation and binarisation techniques was performed. In order to store a feature set in a privacy-preserving manner, it is bound to a secret polynomial in an unlinkable improved fuzzy vault. At authentication, a sufficiently similar feature set will enable the reconstruction of said polynomial where different reconstruction techniques were considered. That is, comparison between the feature sets can be performed while the stored  feature set is permanently protected. Beyond privacy protection the system offers key derivation which can be an essential use-case, \eg for the management of passwords and private keys with biometric characteristics.

In experiments, the proposed system was applied to features extracted with a state-of-the-art deep face recognition system. In cross-database experiments on the FERET and FRGCv2 face databases a false non-match rate below 1\% was achieved at a false match rate of 0.01\%. The conducted security analysis revealed an average security level of up to approximately 28 bits. It has been shown that obtained biometric performance and security rates significantly outperform those reported in the scientific literature. To the best of the authors' knowledge this work represents the first effective application of a fuzzy vault-based template protection schme in order to protect and derive digital keys from deep face representations.

\section*{Acknowledgements}
\label{sec:acknowledgements}
This research work has been partially funded by the German Federal Ministry of Education and Research and the Hessian Ministry of Higher Education, Research, Science and the Arts within their joint support of the National Research Center for Applied Cybersecurity ATHENE.

\bibliographystyle{IEEEtran}
\bibliography{references}

\end{document}